\documentclass[pmlr,twocolumn,10pt]{jmlr} 

\setcitestyle{numbers,square,comma}

\newcommand{\method}{algorithm\xspace}
\newcommand{\methods}{algorithms\xspace}

\newcommand{\gentasks}{randomly generated\xspace}
\newcommand{\Gentasks}{Randomly generated\xspace}

\newcommand{\clintasks}{clinically meaningful\xspace}
\newcommand{\Clintasks}{Clinically meaningful\xspace}

\newcommand{\nmodels}{twelve\xspace}

\usepackage{placeins}
\usepackage{array}
\usepackage{xspace}

\mlhtrack{proceedings}

\newif\iffinal
\finalfalse  

\iffinal
    \ifmlhneedspmlr
      \jmlrvolume{XXX}
      \jmlryear{2026}
    \fi
    \ifmlhfindings \jmlrproceedings{}{ML4H 2026 - Findings Track}\fi
    \ifmlhdemo     \jmlrproceedings{}{ML4H 2026 - Demo Track}\fi
    \jmlrworkshop{Machine Learning for Health (ML4H) 2026}
\else
    \jmlrproceedings{}{Submitted to ML4H 2026: \mlhtrackname}
    \jmlrworkshop{Machine Learning for Health (ML4H) 2026}
\fi

\usepackage{longtable}

\usepackage{booktabs}
\usepackage{multirow}
\usepackage{siunitx}

\usepackage[switch]{lineno}

\theorembodyfont{\upshape}
\theoremheaderfont{\scshape}
\theorempostheader{:}
\theoremsep{\newline}

\title[Methodological Progress Evaluation]{Rethinking How We Evaluate Methodological Progress in Health AI}

\author{%
\Name{Florent Pollet} \Email{ffp2106@cumc.columbia.edu}\and
\Name{Matthew McDermott} \Email{mm6677@cumc.columbia.edu}\\
\addr Department of Biomedical Informatics, Columbia University
}

\begin{document}

\maketitle

\ifmlhdemo\else
\begin{abstract}

Methodological progress in artificial intelligence (AI) for electronic health
records (EHRs) depends on determining which \methods work better, and under
which conditions. However, such progress is thought to be hindered by
difficulties in reproducibility and in defining clinically meaningful evaluation
tasks. We empirically study these barriers by re-implementing 12 historical
and recent \methods within a shared evaluation framework and evaluating them on
two clinical datasets, MIMIC-IV and NWICU. We compare two complementary task
families: expert-authored clinically meaningful tasks and generated tasks defined from randomly sampled
event codes and prediction horizons. We ask whether relative \method comparisons
transfer across task families and datasets, whether residual task heterogeneity
contains useful methodological structure, and what a controlled comparison
reveals about progress over the last decade. Aggregate pairwise comparisons
transfer strongly across evaluation settings, including from randomly generated
to clinically meaningful tasks and across datasets. At the same time, clinically meaningful tasks exhibit
greater task-\method interaction, providing preliminary evidence that task
properties can help explain when particular modeling choices are advantageous.
Finally, newer \methods do not consistently outperform earlier approaches:
gradient-boosted trees remain highly competitive when paired with a high-capacity,
wide and sparse representation of the EHR. Together, these results suggest that
useful methodological knowledge may require less task engineering than commonly
assumed, while highlighting the need to understand the structured heterogeneity
that remains across tasks and \methods.

\end{abstract}
\begin{keywords}
EHR, Evaluation, Health AI
\end{keywords}
\fi

\ifmlhneedsstatements
\paragraph*{Data and Code Availability}

We use the MIMIC-IV and NWICU datasets \citep{PhysioNet-mimiciv-3.1, PhysioNet-nwicu-northwestern-icu-0.1.0}, available on PhysioNet. The code used to reproduce our experiments will be available upon publication.

\paragraph*{Institutional Review Board (IRB)}
This study did not require IRB approval because it involved secondary analysis of existing de-identified datasets and did not involve direct interaction with human participants or access to identifiable private information.
\fi


\section{Introduction}
\label{sec:intro}

Artificial Intelligence (AI) over structured EHR data promises to revolutionize patient care
\citep{topol2019highperformance, boussina2024impact, shimabukuro2017effect, tomasev2019clinically} and is receiving increasing attention in both
the scientific community and the media \citep{wiens2019donoharm,nagendran2020artificial}. Despite this, many researchers remain concerned about
the extent of methodological progress and rigor in this sub-field \citep{mcdermott2025lackscience,bellamy2020evaluating}, arguing that the wealth of research in this space has not meaningfully enabled us to answer key methodological questions, such as when and why some \methods work better than others on a given dataset and task \citep{hardt2022patterns}. 
Commentators have identified many possible barriers to rigorous methodological progress, including:
(i) the difficulty of defining sufficiently \clintasks evaluation tasks to characterize AI \methods \citep{bellamy2020evaluating, mullainathan2022nightingale} and
(ii) the longstanding reproducibility and transportability crisis that hinders replicating task definitions or re-using model code across studies \citep{mcdermott2021reproducibility, johnson2017reproducibility,mcdermott2025lackscience}.

While there is empirical evidence that the reproducibility crisis hinders the accumulation of methodological knowledge--for example, \citeauthor{johnson2017reproducibility} found that the reported performance gain of neural network \methods over gradient-boosted trees often reversed upon reproduction \cite{johnson2017reproducibility}--barrier (i) has received very limited empirical characterization. Concretely, we do not know whether relative comparisons between \methods (e.g., does \method 1 outperform \method 2 when trained from scratch on a given dataset and task) identified on clinically ``meaningless'' tasks would agree with the same comparisons on high-quality, expert-determined tasks. If the conclusions drawn in these two settings were highly similar--and a growing body of evidence on evaluation in other domains of machine learning suggests this is plausible~\citep{salaudeen2026imagenot}--methodological progress would be more accessible in this domain than previously expected, opening new routes to robust insight into the relative performance of different \methods across diverse clinical settings.

In this work, we empirically study this question, and methodological progress in health AI more generally, via a shared implementation
and evaluation framework. We re-implement\footnote{As our models are reproductions, our findings do not imply that the original authors' results were invalid; our reproduction of any given model may itself have errors or flaws. Authors of re-implemented models were not consulted.} historical and recent ICU and EHR
models using a common MEDS-based interface
\citep{mcdermott2026meds}, and evaluate them on two public clinical datasets,
MIMIC-IV \citep{PhysioNet-mimiciv-3.1} and NWICU
\citep{PhysioNet-nwicu-northwestern-icu-0.1.0}, across both \clintasks tasks--explicitly authored by experts to capture meaningful predictions--and \gentasks tasks--which ask whether a randomly sampled code from the dataset vocabulary will occur within a randomly sampled duration, and require no clinical curation. 
We use our findings to answer three key questions:
\textbf{Q1:} Are rankings and comparisons between \methods consistent across evaluation settings, such as across datasets or between \clintasks and \gentasks tasks?
\textbf{Q2:} Beyond global ranking consistency, is there evidence of shared structure between task, \method, and dataset properties that could tell us when a given \method is likely to work well in a target setting?
\textbf{Q3:} When \methods developed over the last decade are evaluated in a common implementation, training, and evaluation framework, what evidence of methodological progress emerges, if any?

Our analyses suggest that:
(1) comparative conclusions about \method performance are statistically significantly similar across evaluation settings, particularly between \clintasks and \gentasks tasks, suggesting that barrier (i) may not be a major barrier for methodological research;
(2) there is preliminary but meaningful evidence of shared structure between tasks, datasets, and \methods, \emph{particularly on \clintasks tasks}, that is suggestive of future opportunities for progress;
and (3) there is meaningful evidence of methodological progress, though not in directions aligned with typical AI development: the best-performing \method in our comparison is a gradient-boosted tree model, but only when paired with an extremely wide, sparse tabularization of the full feature space rather than the restricted, clinically motivated featurizations that dominated historical uses of such models.

In sum, we make the following key contributions:
\begin{enumerate}
    \item We provide reproducible, standard-interface reimplementations of \nmodels AI \methods that can be applied across MEDS datasets through a common interface, enabling robust baseline comparisons in future methodological research.

    \item We provide a rich set of comparisons of \method performance, in both discriminative and computational terms, across a battery of tasks and two datasets, offering concrete guidance on which \methods are likely to perform best and at what computational cost.

    \item We show that \gentasks tasks provide highly reliable estimates of relative \method performance, even across datasets, suggesting a more accessible route to robust methodological research in EHR AI.

    \item We find preliminary evidence that relating task properties to \method and dataset properties yields further methodological insight, suggesting that relative \method rankings are partly predictable from characteristics of the evaluation setting.
\end{enumerate}

\section{Methods}
\label{sec:methods}

To answer the three key questions introduced in Section~\ref{sec:intro}, we use
the following high-level experimental process. First, we re-implement a variety
of historically published \methods such that they are trainable for any binary
classification task on any MEDS dataset through a consistent interface, but are
otherwise faithful to their original, published form. We then train these
aligned \methods on a collection of target tasks from two families:
expert-defined \clintasks tasks and \gentasks tasks, which are randomly defined
and require no clinical expertise. We evaluate the trained \methods by per-task
AUROC on held-out subjects. This process is replicated over two publicly available longitudinal
critical care EHR datasets, MIMIC-IV and NWICU, which differ substantially in
scale and clinical scope (Table~\ref{tab:datasets} in
Appendix~\ref{app:datasets}). With these evaluation results, we answer our
questions directly: for Q1, we compute how consistent the AUROC ranking of these
\methods is across task families and datasets; for Q2, we assess to what extent
variance in \method rankings is related to structured properties of the task,
dataset, and \method combination; and for Q3, we assess how the best overall
performance has changed with the first public artifact date of these \methods and examine
other properties differentiating high- from low-performing \methods. The statistical
procedures used throughout (ranking-agreement tests, task-level reversal rates,
variance decomposition, and uncertainty quantification) are detailed in
Appendix~\ref{app:statistics}.

\subsection{Models}
\label{subsec:models}

\paragraph{Model selection.} Five model families were selected arbitrarily from a set of 49 ICU and EHR modeling approaches identified through a literature review (Appendix~\ref{app:list_models}). 
We deliberately added several models and baselines to anchor the comparison with recent pretraining approaches. MEDS-Tab~\citep{oufattole2024medstab} was included as a strong tabular baseline, using both its XGBoost-based configurations and an additional logistic-regression variant using the same MEDS-Tab featurization (\textsc{MEDS-Tab-LR}). We also included the XGBoost model from the mortality reproducibility study of~\citep{johnson2017reproducibility} (\textsc{ICU-XGBoost}) as a historically motivated ICU baseline. Finally, MOTOR~\citep{steinberg2024motor} and MEDS-EIC-AR~\citep{mcdermott_medseicar} were added as potentially strong recent competitors. Table~\ref{tab:models} summarizes all evaluated model variants.


\paragraph{Model implementation.}
All \nmodels models were re-implemented using agentic generative AI coding workflows (Claude Code with Claude Opus 5; see Appendix~\ref{app:model_development}), with one agent assigned to each model and the implementation template, verification procedure, and human review held fixed across models. Each model therefore received roughly equal ``expert attention'' during implementation, rather than effort concentrating on any single model or family.
To reduce implementation variability and the risk that obvious re-implementation errors affect comparisons, we used a fixed model interface and verification workflow from a common template repository\footnote{Will be released upon publication.
}. For each model, we first inspected the original paper, supplementary material, source code, configurations, preprocessing, and evaluation code, and recorded which components were ported, adapted, or omitted. Implementations then passed a staged set of tests. First, on a small synthetic dataset, we constructed a controlled prediction problem in which a single event code determined a binary label without sequence length or event position leaking the outcome, verified that the model recovered this signal through its actual preprocessing and featurization pipeline, and used shuffled labels as a negative control. Local end-to-end tests also checked that the complete workflow produced predictions for exactly the requested examples and splits. Second, models were executed through the isolated MEDS-DEV interface to verify installation, configuration, and output compatibility. Third, they were run on a MIMIC-IV demo task to expose data-dependent, serialization, resource, and scale-related failures before full-scale experiments. For models requiring specific input features, we supplied dataset-specific predicate files mapping available MEDS events to the required features as closely as possible. Full-data experiments were then run from fixed, clean commits using the same task bundles across models, recording code revision, configuration, dataset, task, and resource information for reproducibility.

\begin{table}[t]
\centering
\caption{Models considered in our evaluation, ordered by the date of their
first publicly available artifact.}
\label{tab:models}
\begin{tabular}{lll}
\toprule
\textbf{Model} & \textbf{Year} & \textbf{Training} \\
\midrule

RETAIN~{\footnotesize \citep{choi2016retain}}
& 2016 & Supervised \\

ICU-XGBoost~{\footnotesize \citep{johnson2017reproducibility}}
& 2017 & Supervised \\

BEHRT~{\footnotesize \citep{li2019behrt}}
& 2019 & Fine-tuned \\

MOTOR-FT~{\footnotesize \citep{steinberg2024motor}}
& 2023 & Fine-tuned \\[-2pt]
{\footnotesize Full fine-tuning (FT)}
& & \\

MOTOR-LP~{\footnotesize \citep{steinberg2024motor}}
& 2023 & Probed \\[-2pt]
{\footnotesize Linear probing (LP)}
& & \\

DuETT~{\footnotesize \citep{labach2023duett}}
& 2023 & Fine-tuned \\

MEDS-TAB-tiny~{\footnotesize \citep{oufattole2024medstab}}
& 2024 & Supervised \\

MEDS-TAB-large~{\footnotesize \citep{oufattole2024medstab}}
& 2024 & Supervised \\

MEDS-TAB-LR~{\footnotesize \citep{oufattole2024medstab}}
& 2024 & Supervised \\

TECO~{\footnotesize \citep{rong2025teco}}
& 2025 & Supervised \\

MEDS-EIC-AR~{\footnotesize \citep{mcdermott_medseicar}}
& 2025 & Fine-tuned \\[-2pt]
{\footnotesize EIC-AR with fine-tuning}
& & \\

MEDS-EIC-AR-sup~{\footnotesize \citep{mcdermott_medseicar}}
& 2025 & Supervised \\[-2pt]
{\footnotesize EIC-AR trained w/o pretraining}
& & \\

\bottomrule
\end{tabular}
\end{table}

\subsection{Tasks}
\label{subsec:tasks}

All tasks are binary prediction problems. For each dataset we use 10 \clintasks
and 10 \gentasks tasks, yielding four evaluation settings (dataset $\times$ task
family) across which the evaluated \methods and evaluation procedure are held
fixed. Full task definitions, cohort sizes, prevalences, and dataset-specific
differences are provided in Appendix~\ref{app:tasks}.

\paragraph{\Gentasks tasks.}
\Gentasks tasks are defined by an event code and a prediction horizon,
inspired by the EveryQuery task parametrization~\citep{chandak2026everyquery}.
For each task, prediction times are sampled from eligible patient timelines, and
the label indicates whether the event occurs within the horizon after the
prediction time. Candidate tasks are retained only when they satisfy
statistical adequacy and censoring requirements; in particular, we require
$\min(n_{\mathrm{pos}}, n_{\mathrm{neg}}) \geq 150$.\footnote{
See Appendix~\ref{app:task-floor} for the derivation of the minority-class
floor and its relationship to AUROC estimation error.
}
Among eligible event-code--horizon pairs, tasks are sampled subject to diversity
constraints limiting repeated codes and over-representation of individual event
categories. Ten tasks are sampled independently per dataset, so the MIMIC-IV
and NWICU \gentasks task sets are intentionally different.

\paragraph{\Clintasks tasks.}
In parallel, we define clinically motivated tasks using ACES
\citep{xu2025aces}, with each task specified by a clinical trigger,
prediction-time anchor, prediction horizon, and, when appropriate, an
exclusion criterion for patients already in the target state. We sought to
make the MIMIC-IV and NWICU suites as similar as possible, adapting task
definitions across datasets when the required information was available, and to
cover diverse clinical specialties, prediction horizons, and question types,
including laboratory abnormalities, acute organ dysfunction, interventions,
utilization, and mortality. Exact correspondence was not required for our main
methodological question, however, and some tasks necessarily differ because the
available events and predicates differ across datasets; we therefore retain
dataset-specific \clintasks tasks when a direct adaptation is not reliable.
Because these more constrained definitions often yield smaller cohorts, we use
a reporting floor of $\min(n_{\mathrm{pos}}, n_{\mathrm{neg}}) \geq 40$ and
exclude tasks below it.

\section{Results}
\label{sec:results}

\subsection{Q1: Comparative conclusions generally transfer across settings}
\label{ssec:q1}

\begin{figure*}[t]
  \centering
  \includegraphics[width=\textwidth]{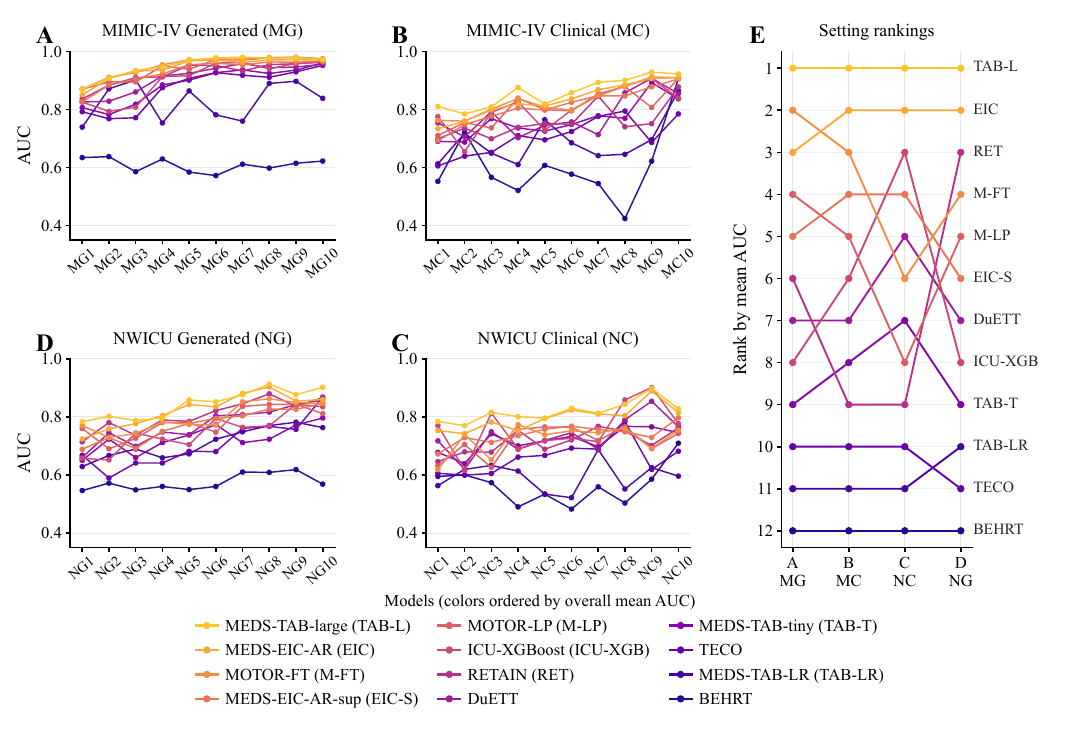}
  \caption{Task-level performance and ranking transfer across settings.
  A--D: tasks ordered from harder to easier by mean AUC across models.
  E: model ranks by setting-level mean AUC (1 = best), ordered MG, MC, NC, NG.
  Colors identify the same models throughout; lines connect tasks or settings
  as visual guides.}
  \label{fig:q1d-task-performance-rankings}
\end{figure*}

Our experiments reliably show that comparative conclusions across \methods
largely transfer from \gentasks tasks to \clintasks tasks and even across
datasets, suggesting that intensive efforts to ensure that benchmarking tasks
are highly clinically meaningful may not be as essential for methodological
research as previously thought.
Figure~\ref{fig:q1d-task-performance-rankings} shows this visually: panels
A--D plot the held-out AUROC of each \method on every task in each of our four
evaluation settings, with \methods colored by global macro-AUROC on a
consistent gradient, so the preservation of this gradient across tasks and
settings shows directly that \method rankings are quite consistent. Full
numerical results are available in Appendix~\ref{app:detailed_results}.

\begin{figure}[t]
  \centering
  \includegraphics[width=0.8\columnwidth]{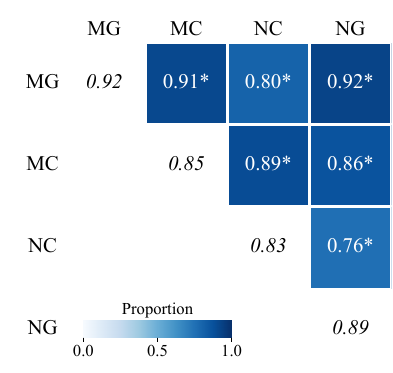}
  \caption{Ranking consistency across evaluation settings. Off-diagonal cells show
  agreement between mean-AUROC \method rankings; italic diagonal entries show one minus
  the within-setting task reversal rate. Stars mark above-chance agreement
  (one-sided random-ranking tests, Holm-adjusted $p<0.05$ across six setting pairs).
  MG/MC denote MIMIC-IV Generated/Clinical; NC/NG denote NWICU Clinical/Generated. Confidence intervals and Kendall $\tau$ for comparisons against MG are available in Appendix \ref{app:new_results}.}
  \label{fig:q1e-consistency-matrix}
\end{figure}

To quantify this finding formally, Figure~\ref{fig:q1e-consistency-matrix}
shows the ranking agreement between every pair of evaluation settings
(Appendix~\ref{app:statistics}). Agreement ranges from 76\% to 92\% and is
statistically significantly higher than chance in all six comparisons after
Holm correction; it is strongest between MIMIC-IV and NWICU \gentasks tasks
(92\%) and between MIMIC-IV \gentasks and \clintasks tasks (91\%).
Figure~\ref{fig:q1d-task-performance-rankings}E further shows that agreement
is concentrated at the top and bottom of the ranking, with more variability in
the middle: MEDS-Tab-large ranks first in every setting, MEDS-EIC-AR ranks second in
three settings and third in the fourth by less than 0.008 AUROC, and the same
three \methods (TECO, MEDS-Tab-LR, and BEHRT) form the bottom three in every
setting.

These results suggest that \gentasks tasks, which require no clinical curation,
can serve as an efficient first-pass substrate for determining which modeling
choices tend to perform better, and that additional datasets need not yield
entirely new comparative conclusions. \Clintasks tasks nonetheless remain
essential when the question shifts from broad \method comparison toward
task-specific or deployment-oriented
evaluation~\citep{salaudeen2025methods,zhang2022shifting}, and with only two
datasets our cross-dataset conclusions remain preliminary.

\subsection{Q2: Dataset-task-\method interactions provide preliminary evidence of shared structure}

While Section~\ref{ssec:q1} shows that comparative conclusions are globally
robust across settings, our experiments are also strongly suggestive of a
shared, underlying structure in task-, dataset-, and \method-performance
relationships. This structure matters because it is precisely the kind of
methodological knowledge we seek to uncover: not only which \method performs
best on average, but which design choices are advantageous under which
conditions. Our analyses find preliminary evidence of this shared structure
through two vehicles: greater \method--task interaction structure among
\clintasks than \gentasks tasks, and relationships between \method design and
task properties that are associated with comparative performance.

\paragraph{Increased heterogeneity and \method--task interaction structure.}
We find three complementary, interrelated sources of evidence suggesting that
the heterogeneity remaining in \method rankings across tasks is driven by
\method--task relationships, especially for \clintasks tasks. First, and most
directly, the within-setting ranking consistency shown on the diagonal of
Figure~\ref{fig:q1e-consistency-matrix} (the proportion of task-level pairwise
\method comparisons that agree with the setting-level aggregate comparison) is
consistently lower for \clintasks tasks than for \gentasks tasks: 15.2\% and
16.7\% of task-level comparisons reverse the aggregate ordering on MIMIC-IV and
NWICU \clintasks tasks, versus 8.3\% and 10.9\% on \gentasks tasks. Second, a
variance decomposition (Appendix~\ref{app:new_results}) estimating how much
observed AUROC variance is due to \methods alone, tasks alone, or their
interaction attributes roughly 24\% of variance to interaction terms for
\clintasks tasks, versus below 10\% for \gentasks tasks. Third, comparing
per-task \method rankings between MIMIC-IV and NWICU across six clinical task
concepts present in both datasets (Appendix~\ref{app:matched_tasks}), rankings
are statistically significantly more similar when the task has the same
conceptual meaning in both datasets than when it does not (mean Kendall's
$\tau_b$ of 0.601 versus 0.437; exact permutation $p=0.019$), suggesting that a
task's conceptual identity is associated with systematic \method-performance
differences that persist across datasets, beyond properties of the \method or
dataset alone.

\paragraph{Evidence of relationships between \method design principles and task properties.}
We can also test whether hypothesized links between \method design principles
and performance hold in real experiments. Specifically, we consider whether the
form of the pre-training loss in our two highest-performing neural \methods
(fine-tuned MOTOR and MEDS-EIC-AR) is related to their relative performance.
MEDS-EIC-AR is pre-trained for next-token prediction, by design a very
short-horizon task, whereas MOTOR is pre-trained to estimate
time-to-next-occurrence distributions over a panel of codes, a much
longer-horizon task. We might therefore hypothesize that MOTOR-FT's advantage
over MEDS-EIC-AR should grow with the task's prediction horizon, which pushes
the task farther from MEDS-EIC-AR's pre-training loss relative to MOTOR's.
Across \gentasks tasks on our two datasets,\footnote{We use only \gentasks
tasks because prediction horizon is an explicitly sampled task parameter in
this family, providing broad horizon variation without tying each horizon to a
clinically authored endpoint.} we find a
statistically significant relationship supporting this hypothesis on NWICU
(Spearman $\rho=0.795$, $p_{\mathrm{Holm}}=0.016$) and a non-significant but
directionally aligned correlation on MIMIC-IV ($\rho=0.483$,
$p_{\mathrm{Holm}}=0.16$), providing preliminary, partial evidence. A similar
hypothesis applies to the underperformance of BEHRT, which by design ingests
only relatively low-frequency input features such as diagnoses and should
therefore perform better on longer-horizon tasks, where such features may play
a more dominant role. We observe a statistically significant relationship
consistent with this hypothesis on MIMIC-IV ($\rho=0.805$,
$p_{\mathrm{Holm}}=0.015$) and a non-significant correlation on NWICU
($\rho=-0.135$, $p_{\mathrm{Holm}}=0.71$). Full tables for these and
further exploratory analyses are available in Appendix~\ref{app:new_results}.

\subsection{Q3: Methodological progress can arise from new uses of existing techniques, not only new architectures}

Our controlled comparison shows mixed evidence of methodological progress over
the last decade, and that progress is poorly summarized by the recency of model
architectures. Most strikingly, the highest-performing pipeline we evaluate uses
gradient-boosted trees, a technology available for roughly the entire period we
study, but obtains its performance through a substantially newer way of
representing and exposing longitudinal EHR data to that learner.

\begin{figure}[t]
  \centering
  \includegraphics[width=\columnwidth]{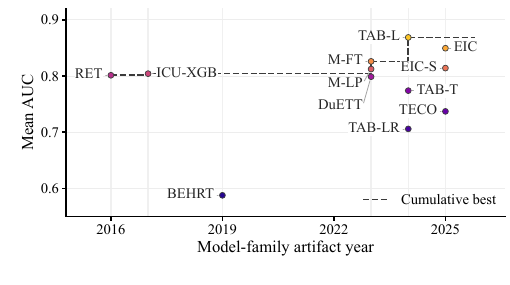}
  \caption{Retrospective progress by model-family artifact year. Each point is
  one model's mean AUROC across the 40 tasks in all four settings; the dashed
  step curve marks the best mean AUROC among models available in that year or
  earlier. Colors and abbreviations match
  Figure~\ref{fig:q1d-task-performance-rankings}. Artifact
  years denote first public availability of the model family; scores are from
  the present evaluation and do not establish historical state of the art.}
  \label{fig:q3-artifact-progress}
\end{figure}

Concretely, Figure~\ref{fig:q3-artifact-progress} shows each \method's mean
AUROC against the date at which, to the best of our knowledge, it was first
released or published. MEDS-Tab-large, an XGBoost-derived
\method~\citep{chen2016xgboost}, obtains best-in-class performance but is
anchored to 2024, when MEDS-Tab was first released. This seems at first glance
contradictory, given that XGBoost models were in use in this space well before
2024; however, MEDS-Tab-large uses XGBoost in a much higher-capacity manner than
many prior XGBoost baselines. For example, ICU-XGBoost, which substantially
underperforms both MEDS-Tab-large and recent neural \methods, uses a more
historically typical formulation: a relatively small set of clinically curated
features modeled through XGBoost. In contrast, MEDS-Tab-large tabularizes the
raw clinical data at an extremely wide, sparse scale, producing millions of features that summarize every code in the dataset vocabulary
over diverse aggregation functions and lookback windows (5,983,579 for MIMIC-IV and 1,103,767 for NWICU). This added
representational capacity, coupled with the advantages of the XGBoost model
class, yields its performance relative to other \methods.

Consistent with this usage, MEDS-Tab-large was among the most time-intensive
\methods to train, as seen in Appendix~\ref{app:compute_tradeoff},
Figure~\ref{fig:q3c-compute-progress-pooled}, which shows the Pareto frontier
between total runtime (preprocessing, amortized pre-training where applicable,
and training) and overall performance. There, MEDS-Tab-large occupies the end
of the frontier that maximizes performance at high cost, while, contrary to
typical expectations, neural \methods such as MEDS-EIC-AR offer competitive
performance at greatly reduced wall-time cost, a benefit to efficiency rather
than predictive power.

\section{Discussion}
\label{sec:discussion}

\subsection{Implications for methodological research in Health AI}

Taken together, Q1 and Q2 suggest a two-stage evaluation strategy for
methodological research in Health AI: first estimate the broadly stable
component of \method performance, which transfers across task families and
datasets and can therefore be measured cheaply, and then study systematic
departures from it, which is where task-, dataset-, and design-specific
methodological knowledge lives. Under this view, the question to ask of a new
\method is not only whether it improves a particular benchmark, but whether
that improvement transfers, where it fails, and what those failures teach us
about which \methods work where.

A natural extension is to characterize the task space itself. With more tasks
and datasets, one could ask whether subsets of tasks induce consistently similar
\method rankings, whether such subsets can be predicted from observable
properties such as horizon, prevalence, clinical category, or cohort size, and
whether \method characteristics explain systematic departures from the
aggregate ranking. Such analyses could turn task heterogeneity from a nuisance
into a source of methodological insight.

Finally, our experience suggests that recent tooling can considerably lower the
historical barriers to comparative research in Health
AI~\citep{mcdermott2021reproducibility,donoho2024data}. Agentic generative AI
made it feasible to reconstruct multiple heterogeneous model pipelines under a
common interface with comparable effort per model, while data standards and
shared tooling---MEDS for event-stream representation~\citep{mcdermott2026meds},
MEDS-DEV for decentralized model execution and comparison~\citep{kolo2024medsdev},
and ACES for portable task specification~\citep{xu2025aces}---make such
comparisons cumulative rather than one-off. Standardized re-implementation is a
different objective from exact reproduction of an original result, and our
framework is designed for the former. We intend to contribute the re-implemented
\methods back to MEDS-DEV so that future work can extend rather than recreate
these comparisons.

\subsection{Limitations and future work}

Several limitations constrain the scope of our conclusions. First, we evaluate
only two datasets, both drawn from structured hospital and ICU EHR data;
extending the study to additional public datasets such as eICU and EHRSHOT, and
to substantially different clinical contexts, would provide a stronger test of
dataset invariance. Second, our \method set is a subset of the much larger
Health AI literature, and extending the common interface to the remaining
candidate \methods would reduce sensitivity to the particular \methods sampled
here. Third, each evaluation setting contains only ten tasks, which is
sufficient to identify aggregate transfer and task-level heterogeneity but too
small to characterize a latent task space; multiple independently sampled
\gentasks suites and larger \clintasks collections would enable sensitivity
analyses with respect to task count. Fourth, \clintasks task definitions cannot
always be transferred identically across datasets, so some observed
cross-dataset differences may conflate dataset and task-definition changes. Fifth, all re-implementations are best-effort
translations of the original \methods: our common template, behavioral tests,
integration tests, and human review reduce the risk of obvious implementation
errors but cannot guarantee equivalence to the original pipelines, so our
conclusions concern the \methods as instantiated within our shared framework
rather than the original studies. Finally, our uncertainty analysis does not
capture all sources of variation: we do not systematically repeat model
training across random seeds, and our primary analyses focus on AUROC rather
than other metrics such as win rate.

\section{Related work}
\label{sec:related_work}

Prior work on barriers to cumulative methodological progress in Health AI can
be organized around three recurring themes: access to and suitability of
clinical data \citep{mullainathan2022nightingale,zhang2022shifting}, the formulation of meaningful and comparable evaluation tasks \citep{johnson2017reproducibility,bellamy2020evaluating},
and the reproducibility and comparability of experimental results \citep{mcdermott2021reproducibility}. 
Donoho's account of frictionless reproducibility identifies complementary
ingredients for such cumulative comparison: accessible data,
re-executable workflows, and shared challenge problems with explicit
performance criteria~\citep{donoho2024data}. Shared datasets alone, however,
do not guarantee comparability. In a mortality-prediction case study,
\citet{johnson2017reproducibility} documented substantial heterogeneity in
cohort construction and task specification, motivating shared code, benchmarks,
and common data-extraction procedures. \citet{harutyunyan2019multitask}
subsequently introduced four standardized clinical prediction tasks on
MIMIC-III, while MIMIC-Extract~\citep{wang2020mimicextract} further
standardized data extraction and preprocessing for reproducible EHR modeling.
Yet \citet{bellamy2020evaluating} found that only a small fraction of
subsequent studies citing these benchmarks used sufficiently similar
experimental setups for direct comparison, so benchmark availability and
citation alone do not ensure consistent use.

More recent benchmarks have expanded both the scope of tasks and the models
being evaluated. EHRSHOT~\citep{wornow2023ehrshot} defines 15 prediction tasks
over longitudinal EHR data for few-shot evaluation of pretrained models, moving
beyond ICU-only evaluation. FoMoH~\citep{pang2025fomoh} similarly evaluates
structured-EHR foundation models across a diverse suite of clinically meaningful
prediction tasks, emphasizing clinical relevance and cross-model comparability. These efforts provide increasingly rich and
standardized evaluation settings, but necessarily instantiate particular
choices of datasets, cohorts, and tasks. Our work is complementary: rather
than proposing another fixed benchmark, we ask how much the methodological
conclusions obtained from an evaluation depend on those choices themselves.





\FloatBarrier

\bibliography{ref}

@article{PhysioNet-mimiciv-3.1,
  author = {Johnson, Alistair and Bulgarelli, Lucas and Pollard, Tom and Gow, Brian and Moody, Benjamin and Horng, Steven and Celi, Leo Anthony and Mark, Roger},
  title = {{MIMIC-IV}},
  journal = {{PhysioNet}},
  year = {2024},
  month = oct,
  note = {Version 3.1},
  doi = {10.13026/kpb9-mt58},
  url = {https://doi.org/10.13026/kpb9-mt58}
}

@article{PhysioNet-nwicu-northwestern-icu-0.1.0,
  author = {Moukheiber, Dana and Temps, William and Molgi, Bhadrappa and Li, Yikuan and Lu, Alice and Nannapaneni, Prasanth and Chahin, Abdulrahman and Hao, Sicheng and {Torres Fabregas}, Felipe and Celi, Leo Anthony and Wong, Adrian and Lloyd, Maxwell and {Borrat Frigola}, Xavier and Lee, Hyung-Chul and Schneider, Daniel and Pollard, Tom and Luo, Yuan and Kho, Abel and Mark, Roger},
  title = {{Northwestern ICU (NWICU) database}},
  journal = {{PhysioNet}},
  year = {2024},
  month = nov,
  note = {Version 0.1.0},
  doi = {10.13026/s84w-1829},
  url = {https://doi.org/10.13026/s84w-1829}
}

@article{topol2019highperformance,
  author  = {Topol, Eric J.},
  title   = {High-performance medicine: the convergence of human and artificial intelligence},
  journal = {Nature Medicine},
  year    = {2019},
  volume  = {25},
  number  = {1},
  pages   = {44--56},
  doi     = {10.1038/s41591-018-0300-7}
}

@article{wiens2019donoharm,
  author  = {Wiens, Jenna and Saria, Suchi and Sendak, Mark and Ghassemi, Marzyeh
             and Liu, Vincent X. and Doshi-Velez, Finale and Jung, Kenneth
             and Heller, Katherine and Kale, David and Saeed, Mohammed
             and Ossorio, Pilar N. and Thadaney-Israni, Sonoo
             and Goldenberg, Anna},
  title   = {Do no harm: a roadmap for responsible machine learning for health care},
  journal = {Nature Medicine},
  year    = {2019},
  volume  = {25},
  number  = {9},
  pages   = {1337--1340},
  doi     = {10.1038/s41591-019-0548-6}
}

@article{nagendran2020artificial,
  author  = {Nagendran, Myura and Chen, Yang and Lovejoy, Christopher A.
             and Gordon, Anthony C. and Komorowski, Matthieu and Harvey, Hugh
             and Topol, Eric J. and Ioannidis, John P. A.
             and Collins, Gary S. and Maruthappu, Mahiben},
  title   = {Artificial intelligence versus clinicians: systematic review of
             design, reporting standards, and claims of deep learning studies},
  journal = {BMJ},
  year    = {2020},
  volume  = {368},
  pages   = {m689},
  doi     = {10.1136/bmj.m689}
}

@inproceedings{mcdermott2025lackscience,
  title     = {The (lack of?) Science of Machine Learning for Healthcare},
  author    = {McDermott, Matthew},
  booktitle = {Proceedings of the 4th Machine Learning for Health Symposium},
  pages     = {19--29},
  year      = {2025},
  editor    = {Hegselmann, Stefan and Zhou, Helen and Healey, Elizabeth
               and Chang, Trenton and Ellington, Caleb and Mhasawade, Vishwali
               and Tonekaboni, Sana and Argaw, Peniel and Zhang, Haoran},
  volume    = {259},
  series    = {Proceedings of Machine Learning Research},
  publisher = {PMLR},
  url       = {https://proceedings.mlr.press/v259/mcdermott25a.html}
}

@article{bellamy2020evaluating,
  title   = {Evaluating Progress on Machine Learning for Longitudinal Electronic Healthcare Data},
  author  = {Bellamy, David and Celi, Leo and Beam, Andrew L.},
  journal = {arXiv preprint arXiv:2010.01149},
  year    = {2020},
  url     = {https://arxiv.org/abs/2010.01149}
}

@book{hardt2022patterns,
  title     = {Patterns, Predictions, and Actions: Foundations of Machine Learning},
  author    = {Hardt, Moritz and Recht, Benjamin},
  year      = {2022},
  publisher = {Princeton University Press},
  isbn      = {9780691233734}
}

@inproceedings{salaudeen2025methods,
  title     = {On Evaluating Methods vs. Evaluating Models},
  author    = {Salaudeen, Olawale and Dorner, Florian E. and Hase, Peter},
  booktitle = {NeurIPS 2025 Workshop on Evaluating the Evolving LLM Lifecycle:
               Benchmarks, Emergent Abilities, and Scaling},
  year      = {2025}
}

@article{zhang2022shifting,
  title   = {Shifting machine learning for healthcare from development to
             deployment and from models to data},
  author  = {Zhang, Angela and Xing, Lei and Zou, James and Wu, Joseph C.},
  journal = {Nature Biomedical Engineering},
  volume  = {6},
  pages   = {1330--1345},
  year    = {2022},
  doi     = {10.1038/s41551-022-00898-y}
}

@article{mcdermott2021reproducibility,
  title   = {Reproducibility in machine learning for health research: Still a ways to go},
  author  = {McDermott, Matthew B. A. and Wang, Shirly and Marinsek, Nikolay
             and Ranganath, Rajesh and Foschini, Luca and Ghassemi, Marzyeh},
  journal = {Science Translational Medicine},
  volume  = {13},
  number  = {586},
  pages   = {eabb1655},
  year    = {2021},
  doi     = {10.1126/scitranslmed.abb1655}
}

@article{mullainathan2022nightingale,
  author  = {Mullainathan, Sendhil and Obermeyer, Ziad},
  title   = {Solving medicine's data bottleneck: Nightingale Open Science},
  journal = {Nature Medicine},
  year    = {2022},
  month   = may,
  volume  = {28},
  number  = {5},
  pages   = {897--899},
  doi     = {10.1038/s41591-022-01804-4},
  url     = {https://doi.org/10.1038/s41591-022-01804-4}
}

@inproceedings{johnson2017reproducibility,
  title     = {Reproducibility in critical care: a mortality prediction case study},
  author    = {Johnson, Alistair E. W. and Pollard, Tom J. and Mark, Roger G.},
  booktitle = {Proceedings of the 2nd Machine Learning for Healthcare Conference},
  pages     = {361--376},
  year      = {2017},
  editor    = {Doshi-Velez, Finale and Fackler, Jim and Kale, David
               and Ranganath, Rajesh and Wallace, Byron and Wiens, Jenna},
  volume    = {68},
  series    = {Proceedings of Machine Learning Research},
  publisher = {PMLR},
  url       = {https://proceedings.mlr.press/v68/johnson17a.html}
}

@article{mcdermott2026meds,
  title   = {{MEDS}---An Emerging Data Standard and Ecosystem for Health AI Research},
  author  = {McDermott, Matthew B. A. and Steinberg, Ethan and Fries, Jason A.
             and van de Water, Robin P. and Pang, Chao and Rockenschaub, Patrick
             and Renc, Pawel and Oh, Jungwoo and Stankevi{\v{c}}i{\=u}t{\.e}, Kamil{\.e}
             and Xu, Justin and Pollard, Tom J. and Oufattole, Nassim
             and Wornow, Michael and Bergamaschi, Teya S. and Jeong, Hyewon
             and Lee, Simon A. and Jeanselme, Vincent and Klein, Kiril V.
             and Odgaard, Mikkel and Montgomery, Maria E. and Sitek, Arkadiusz
             and Nielsen, Mads and Chiang, Jeffrey N. and Dagan, Noa
             and Kohane, Isaac and Joshi, Shalmali and Choi, Edward
             and Shah, Nigam H.},
  journal = {NEJM AI},
  volume  = {3},
  number  = {6},
  year    = {2026},
  doi     = {10.1056/AIra2501253}
}

@inproceedings{kolo2024medsdev,
  title     = {{MEDS} Decentralized, Extensible Validation ({MEDS-DEV}) Benchmark:
               Establishing Reproducibility and Comparability in ML for Health},
  author    = {Kolo, Aleksia and Pang, Chao and Choi, Edward and Steinberg, Ethan
               and Jeong, Hyewon and Gallifant, Jack and Fries, Jason A.
               and Chiang, Jeffrey N. and Oh, Jungwoo and Xu, Justin
               and Stankevi{\v{c}}i{\=u}t{\.e}, Kamil{\.e} and Klein, Kiril V.
               and McDermott, Matthew B. A. and Odgaard, Mikkel
               and Oufattole, Nassim and Shah, Nigam H. and Rockenschaub, Patrick
               and Renc, Pawel and van de Water, Robin P. and Joshi, Shalmali
               and Lee, Simon A. and Bergamaschi, Teya S. and Pollard, Tom J.
               and Jeanselme, Vincent and Choi, Young Sang and Wornow, Michael
               and Kashyap, Apara and Jiang, Xinzhuo and Li, Yanwei
               and Kobayashi, Yuta and King, Ryan C.},
  booktitle = {Machine Learning for Health (ML4H) 2024 Demo Track},
  year      = {2024}
}

@inproceedings{chen2016xgboost,
  title     = {{XGBoost}: A Scalable Tree Boosting System},
  author    = {Chen, Tianqi and Guestrin, Carlos},
  booktitle = {Proceedings of the 22nd ACM SIGKDD International Conference on
               Knowledge Discovery and Data Mining},
  pages     = {785--794},
  year      = {2016},
  publisher = {ACM},
  doi       = {10.1145/2939672.2939785}
}

@article{salaudeen2026imagenot,
  title   = {ImageNot: A Contrast with ImageNet Preserves Model Rankings},
  author  = {Salaudeen, Olawale and Hardt, Moritz},
  journal = {Transactions on Machine Learning Research},
  year    = {2026}
}

@misc{oufattole2024medstab,
  title        = {{MEDS-Tab}: Automated tabularization and baseline methods for {MEDS} datasets},
  author       = {Oufattole, Nassim and Bergamaschi, Teya and Kolo, Aleksia
                  and Jeong, Hyewon and Gaggin, Hanna and Stultz, Collin M.
                  and McDermott, Matthew B. A.},
  year         = {2024},
  eprint       = {2411.00200},
  archivePrefix= {arXiv},
  primaryClass = {cs.LG},
  url          = {https://arxiv.org/abs/2411.00200}
}

@inproceedings{steinberg2024motor,
  title     = {{MOTOR}: A Time-to-Event Foundation Model for Structured Medical Records},
  author    = {Steinberg, Ethan and Fries, Jason Alan and Xu, Yizhe and Shah, Nigam},
  booktitle = {International Conference on Learning Representations},
  year      = {2024}
}

@software{mcdermott_medseicar,
  author = {McDermott, Matthew},
  title  = {{MEDS} ``Everything-is-code'' Autoregressive Model},
  doi    = {10.5281/zenodo.17535559},
  url    = {https://github.com/mmcdermott/MEDS_EIC_AR}
}

@article{choi2016retain,
  title   = {{RETAIN}: An Interpretable Predictive Model for Healthcare
             using Reverse Time Attention Mechanism},
  author  = {Choi, Edward and Bahadori, Mohammad Taha and Kulas, Joshua A.
             and Schuetz, Andy and Stewart, Walter F. and Sun, Jimeng},
  journal = {arXiv preprint arXiv:1608.05745},
  year    = {2016}
}

@article{li2019behrt,
  title   = {{BEHRT}: Transformer for Electronic Health Records},
  author  = {Li, Yikuan and Rao, Shishir and Ayala Solares, Jose Roberto
             and Hassaine, Abdelaali and Canoy, Dexter and Zhu, Yajie
             and Rahimi, Kazem and Salimi-Khorshidi, Gholamreza},
  journal = {arXiv preprint arXiv:1907.09538},
  year    = {2019}
}

@article{labach2023duett,
  title   = {{DuETT}: Dual Event Time Transformer for Electronic Health Records},
  author  = {Labach, Alex and Pokhrel, Aslesha and Huang, Xiao Shi
             and Zuberi, Saba and Yi, Seung Eun and Volkovs, Maksims
             and Poutanen, Tomi and Krishnan, Rahul G.},
  journal = {arXiv preprint arXiv:2304.13017},
  year    = {2023}
}

@article{rong2025teco,
  title   = {A Deep Learning Model for Clinical Outcome Prediction Using
             Longitudinal Inpatient Electronic Health Records},
  author  = {Rong, Ruichen and Gu, Zifan and Lai, Hongyin and Nelson, Tanna L.
             and Keller, Tony and Walker, Clark and Jin, Kevin W.
             and Chen, Catherine and Navar, Ann Marie and Velasco, Ferdinand
             and Peterson, Eric D. and Xiao, Guanghua and Yang, Donghan M.
             and Xie, Yang},
  journal = {medRxiv},
  year    = {2025},
  doi     = {10.1101/2025.01.21.25320916}
}

@misc{chandak2026everyquery,
  title        = {EveryQuery: Zero-Shot Clinical Prediction via Task-Conditioned Pretraining over Electronic Health Records},
  author       = {Chandak, Payal and Kondas, Gregory and Antwarg Friedman, Liat
                  and Kohane, Isaac and McDermott, Matthew},
  year         = {2026},
  eprint       = {2603.07900},
  archivePrefix= {arXiv},
  primaryClass = {cs.AI},
  url          = {https://arxiv.org/abs/2603.07900}
}

@inproceedings{xu2025aces,
  title     = {{ACES}: Automatic Cohort Extraction System for Event-Stream Datasets},
  author    = {Xu, Justin and Gallifant, Jack and Johnson, Alistair E. W.
               and McDermott, Matthew B. A.},
  booktitle = {International Conference on Learning Representations},
  year      = {2025},
  url       = {https://proceedings.iclr.cc/paper_files/paper/2025/hash/d8542126cd3e0dd6c0a44e0aa1957072-Abstract-Conference.html}
}

@article{hanley1982meaning,
  title   = {The Meaning and Use of the Area under a Receiver Operating
             Characteristic ({ROC}) Curve},
  author  = {Hanley, James A. and McNeil, Barbara J.},
  journal = {Radiology},
  volume  = {143},
  number  = {1},
  pages   = {29--36},
  year    = {1982},
  doi     = {10.1148/radiology.143.1.7063747}
}

@article{donoho2024data,
  author = {Donoho, David},
  title = {Data Science at the Singularity},
  journal = {Harvard Data Science Review},
  year = {2024},
  volume = {6},
  number = {1},
  doi = {10.1162/99608f92.b91339ef},
  url = {https://hdsr.mitpress.mit.edu/pub/g9mau4m0/release/2}
}

@article{harutyunyan2019multitask,
  author = {Harutyunyan, Hrayr and Khachatrian, Hrant and Kale, David C. and Ver Steeg, Greg and Galstyan, Aram},
  title = {Multitask learning and benchmarking with clinical time series data},
  journal = {Scientific Data},
  year = {2019},
  volume = {6},
  number = {1},
  pages = {96},
  doi = {10.1038/s41597-019-0103-9},
  url = {https://doi.org/10.1038/s41597-019-0103-9}
}

@article{pang2025fomoh,
  title   = {{FoMoH}: A Clinically Meaningful Foundation Model Evaluation for Structured Electronic Health Records},
  author  = {Pang, Chao and Jeanselme, Vincent and Choi, Young Sang
             and Jiang, Xinzhuo and Jing, Zilin and Kashyap, Aparajita
             and Kobayashi, Yuta and Li, Yanwei and Pollet, Florent
             and Natarajan, Karthik and Joshi, Shalmali},
  journal = {arXiv preprint arXiv:2505.16941},
  year    = {2025},
  doi     = {10.48550/arXiv.2505.16941}
}

@inproceedings{wornow2023ehrshot,
  title     = {{EHRSHOT}: An EHR Benchmark for Few-Shot Evaluation of Foundation Models},
  author    = {Wornow, Michael and Thapa, Rahul and Steinberg, Ethan
               and Fries, Jason A. and Shah, Nigam H.},
  booktitle = {Advances in Neural Information Processing Systems},
  volume    = {36},
  pages     = {67125--67137},
  year      = {2023},
  doi       = {10.52202/075280-2933}
}

@inproceedings{wang2020mimicextract,
  title     = {{MIMIC-Extract}: A Data Extraction, Preprocessing, and Representation Pipeline for {MIMIC-III}},
  author    = {Wang, Shirly and McDermott, Matthew B. A. and Chauhan, Geeticka
               and Ghassemi, Marzyeh and Hughes, Michael C. and Naumann, Tristan},
  booktitle = {Proceedings of the ACM Conference on Health, Inference, and Learning},
  pages     = {222--235},
  year      = {2020},
  publisher = {ACM},
  doi       = {10.1145/3368555.3384469}
}

@article{boussina2024impact,
  title     = {Impact of a deep learning sepsis prediction model on quality of care and survival},
  author    = {Boussina, Aaron and Shashikumar, Supreeth P. and Malhotra, Atul and Owens, Robert L. and El-Kareh, Robert and Longhurst, Christopher A. and Quintero, Kimberly and Donahue, Allison and Chan, Theodore C. and Nemati, Shamim and Wardi, Gabriel},
  journal   = {npj Digital Medicine},
  volume    = {7},
  pages     = {14},
  year      = {2024},
  doi       = {10.1038/s41746-023-00986-6}
}

@article{shimabukuro2017effect,
  title     = {Effect of a machine learning-based severe sepsis prediction algorithm on patient survival and hospital length of stay: a randomised clinical trial},
  author    = {Shimabukuro, David W. and Barton, Christopher W. and Feldman, Mitchell D. and Mataraso, Samson J. and Das, Ritankar},
  journal   = {BMJ Open Respiratory Research},
  volume    = {4},
  number    = {1},
  pages     = {e000234},
  year      = {2017},
  doi       = {10.1136/bmjresp-2017-000234},
  pmid      = {29435343},
  pmcid     = {PMC5687546}
}

@article{tomasev2019clinically,
  title     = {A clinically applicable approach to continuous prediction of future acute kidney injury},
  author    = {Toma{\v{s}}ev, Nenad and Glorot, Xavier and Rae, Jack W. and Zielinski, Michal and Askham, Harry and Saraiva, Andre and Mottram, Anne and Meyer, Clemens and Ravuri, Suman and Protsyuk, Ivan and Connell, Alistair and Hughes, C{\'i}an O. and Karthikesalingam, Alan and Cornebise, Julien and Montgomery, Hugh and Rees, Geraint and Laing, Chris and Baker, Clifton R. and Peterson, Kelly and Reeves, Ruth and Hassabis, Demis and King, Dominic and Suleyman, Mustafa and Back, Trevor and Nielson, Christopher and Ledsam, Joseph R. and Mohamed, Shakir},
  journal   = {Nature},
  volume    = {572},
  number    = {7767},
  pages     = {116--119},
  year      = {2019},
  doi       = {10.1038/s41586-019-1390-1},
  pmid      = {31367026},
  pmcid     = {PMC6722431}
}

\appendix

\onecolumn

\section{List of considered models}
\label{app:list_models}

The list of considered ICU/EHR models for our study is presented in Table \ref{tab:model-cards}.

\begingroup
\small
\setlength{\tabcolsep}{5pt}

\begin{longtable}{@{}lll@{}}
\caption{The 49 structured-EHR model cards surveyed for this pilot, grouped by
training strategy. \emph{Supervised} models learn predictors directly from
task-specific labels; \emph{Pretrained} models use a separate reusable
pretraining stage, consumed either by fine-tuning or by a frozen probe;
\emph{Other} includes zero-shot and prior-fitted approaches that do not require
task-specific supervised training. For zero-shot \methods, \emph{dir.} denotes
\methods that produce predictions directly from the observed record without
sampling future trajectories, whereas \emph{mat.} denotes \methods that obtain
predictions by sampling or materializing future trajectories and evaluating the
target outcome on those trajectories. Years follow each model card's
classification; links are to the source paper and, where available, a public
code repository. Years in this table refer to publication year, whereas years in some other tables refer to the artifact date (e.g., the date of the original preprint or release). $^*$ foundation-model configuration.
$^\dagger$ with self-supervised pretraining.}
\label{tab:model-cards}\\

\toprule
\textbf{Model} & \textbf{Publication Year} & \textbf{Resource} \\
\midrule
\endfirsthead

\caption[]{\emph{(continued)}}\\
\toprule
\textbf{Model} & \textbf{Publication Year} & \textbf{Resource} \\
\midrule
\endhead

\midrule
\multicolumn{3}{r@{}}{\emph{continued on next page}}\\
\endfoot

\bottomrule
\endlastfoot

\multicolumn{3}{@{}l}{\textbf{Supervised}}\\
\addlinespace[2pt]

STraTS-mTAND
& 2025
& \href{https://dl.acm.org/doi/10.1145/3743689}{paper} \\

TECO
& 2025
& \href{https://academic.oup.com/jamiaopen/article/8/2/ooaf026/8110091}{paper} \\

GenHPF
& 2024
& \href{https://arxiv.org/abs/2207.09858}{paper} /
  \href{https://github.com/hoon9405/GenHPF}{code} \\

PRISM
& 2024
& \href{https://arxiv.org/abs/2309.04160}{paper} /
  \href{https://github.com/yhzhu99/PRISM}{code} \\

TRANS
& 2024
& \href{https://arxiv.org/abs/2405.03943}{paper} /
  \href{https://github.com/The-Real-JerryChen/TRANS}{code} \\

KerPrint
& 2023
& \href{https://ojs.aaai.org/index.php/AAAI/article/view/25667/25439}{paper} /
  \href{https://github.com/xyxpku/KerPrint}{code} \\

Chet
& 2022
& \href{https://arxiv.org/abs/2112.05195}{paper} /
  \href{https://github.com/LuChang-CS/Chet}{code} \\

UniHPF
& 2022
& \href{https://arxiv.org/abs/2211.08082}{paper} /
  \href{https://github.com/hoon9405/UniHPF}{code} \\

GRASP
& 2021
& \href{https://ojs.aaai.org/index.php/AAAI/article/view/16152}{paper} /
  \href{https://github.com/choczhang/GRASP}{code} \\

SETOR
& 2021
& \href{https://arxiv.org/abs/2109.03069}{paper} /
  \href{https://github.com/Xueping/SETOR}{code} \\

AdaCare
& 2020
& \href{https://arxiv.org/abs/1911.12205}{paper} /
  \href{https://github.com/Accountable-Machine-Intelligence/AdaCare}{code} \\

ConCare
& 2020
& \href{https://arxiv.org/abs/1911.12216}{paper} /
  \href{https://github.com/Accountable-Machine-Intelligence/ConCare}{code} \\

GCT
& 2020
& \href{https://arxiv.org/abs/1906.04716}{paper} /
  \href{https://github.com/lycpaul/dl4h-gp30-gct}{code} \\

HiTANet
& 2020
& \href{https://dl.acm.org/doi/10.1145/3394486.3403107}{paper} /
  \href{https://github.com/machinelearning4health/HiTANet}{code} \\

StageNet
& 2020
& \href{https://arxiv.org/abs/2001.10054}{paper} /
  \href{https://github.com/v1xerunt/StageNet}{code} \\

SAnD
& 2018
& \href{https://arxiv.org/abs/1711.03905}{paper} /
  \href{https://github.com/khirotaka/SAnD}{code} \\

Dipole
& 2017
& \href{https://arxiv.org/abs/1706.05764}{paper} /
  \href{https://github.com/sunlabuiuc/PyHealth}{code} \\

GRAM
& 2017
& \href{https://arxiv.org/abs/1611.07012}{paper} /
  \href{https://github.com/mp2893/gram}{code} \\

Doctor AI
& 2016
& \href{https://proceedings.mlr.press/v56/Choi16.html}{paper} /
  \href{https://github.com/mp2893/doctorai}{code} \\

RETAIN
& 2016
& \href{https://proceedings.neurips.cc/paper/2016/hash/231141b34c82aa95e48810a9d1b33a79-Abstract.html}{paper} /
  \href{https://github.com/mp2893/retain}{code} \\

\addlinespace
\multicolumn{3}{@{}l}{\textbf{Pretrained -- Probe}}\\
\addlinespace[2pt]

ORA
& 2026
& \href{https://arxiv.org/abs/2602.00541}{paper} \\

PORTER
& 2026
& \href{https://arxiv.org/abs/2606.24102}{paper} \\

CLMBR
& 2021
& \href{https://arxiv.org/abs/2001.05295}{paper} /
  \href{https://github.com/som-shahlab/ehr_ml}{code} \\

\addlinespace
\multicolumn{3}{@{}l}{\textbf{Pretrained -- Fine-tuned}}\\
\addlinespace[2pt]

AID-MAE
& 2026
& \href{https://arxiv.org/abs/2602.15159}{paper} \\

HealthFormer
& 2026
& \href{https://www.medrxiv.org/content/10.64898/2026.03.25.26349262v2.full-text}{paper} /
  \href{https://github.com/renyi-ai/HealthFormer}{code} \\

SurvivEHR
& 2026
& \href{https://www.nature.com/articles/s41746-026-02709-z}{paper} /
  \href{https://github.com/cwlgadd/SurvivEHR}{code} \\

BAT$^*$
& 2025
& \href{https://arxiv.org/abs/2509.19885}{paper} /
  \href{https://github.com/Katja-Jagd/YAIB}{code} \\

CEHR-XGPT
& 2025
& \href{https://arxiv.org/abs/2509.03643}{paper} /
  \href{https://github.com/knatarajan-lab/cehrgpt}{code} \\

PULSE-ICU
& 2025
& \href{https://arxiv.org/abs/2511.22199}{paper} /
  \href{https://github.com/sejeongak/PULSE-ICU}{code} \\

TOO-BERT
& 2025
& \href{https://medinform.jmir.org/2025/1/e68138}{paper} /
  \href{https://github.com/ali-amirahmadii/TOO-Bert}{code} \\

CORE-BEHRT
& 2024
& \href{https://raw.githubusercontent.com/mlresearch/v252/main/assets/odgaard24a/odgaard24a.pdf}{paper} /
  \href{https://github.com/mikkelfo/CORE-BEHRT}{code} \\

EBCL
& 2024
& \href{https://proceedings.mlr.press/v252/oufattole24a.html}{paper} /
  \href{https://github.com/mit-ccrg/EBCL}{code} \\

MOTOR
& 2024
& \href{https://proceedings.iclr.cc/paper_files/paper/2024/hash/b36554b97da741b1c48c9de05c73993e-Abstract-Conference.html}{paper} /
  \href{https://github.com/som-shahlab/motor_code_release}{code} \\

DuETT
& 2023
& \href{https://proceedings.mlr.press/v219/labach23a.html}{paper} /
  \href{https://github.com/layer6ai-labs/duett}{code} \\

Hi-BEHRT
& 2023
& \href{https://arxiv.org/abs/2106.11360}{paper} \\

TransEHR
& 2023
& \href{https://proceedings.mlr.press/v225/xu23a.html}{paper} /
  \href{https://github.com/SigmaTsing/TransEHR}{code} \\

TransformEHR
& 2023
& \href{https://www.nature.com/articles/s41467-023-43715-z}{paper} /
  \href{https://github.com/whaleloops/TransformEHR}{code} \\

GenHPF$^\dagger$
& 2022/2023
& \href{https://arxiv.org/abs/2207.09858}{paper} /
  \href{https://github.com/hoon9405/GenHPF}{code} \\

STraTS
& 2022
& \href{https://arxiv.org/abs/2107.14293}{paper} /
  \href{https://github.com/sindhura97/STraTS}{code} \\

CEHR-BERT
& 2021
& \href{https://proceedings.mlr.press/v158/pang21a.html}{paper} /
  \href{https://github.com/cumc-dbmi/cehrbert}{code} \\

Med-BERT
& 2021
& \href{https://www.nature.com/articles/s41746-021-00455-y}{paper} /
  \href{https://github.com/ZhiGroup/Med-BERT}{code} \\

BEHRT
& 2020
& \href{https://www.nature.com/articles/s41598-020-62922-y}{paper} /
  \href{https://github.com/deepmedicine/BEHRT}{code} \\

G-BERT
& 2019
& \href{https://arxiv.org/abs/1906.00346}{paper} /
  \href{https://github.com/jshang123/g-bert}{code} \\

\addlinespace
\multicolumn{3}{@{}l}{\textbf{Other}}\\
\addlinespace[2pt]

EveryQuery (dir.)
& 2026
& \href{https://arxiv.org/abs/2603.07900}{paper} /
  \href{https://github.com/payalchandak/EveryQuery}{code} \\

SurvPFN
& 2026
& \href{https://arxiv.org/abs/2606.04564}{paper} /
  \href{https://github.com/genepi-freiburg/SurvPFN}{code} \\

CoMET (mat.)
& 2025
& \href{https://arxiv.org/abs/2508.12104}{paper} \\

Delphi-2M (dir.)
& 2025
& \href{https://www.nature.com/articles/s41586-025-09529-3}{paper} /
  \href{https://github.com/gerstung-lab/Delphi}{code} \\

MEDS-EIC-AR (mat.)
& 2025--2026
& \href{https://github.com/mmcdermott/MEDS_EIC_AR}{code} \\

ETHOS / ARES (mat.)
& 2024/2025
& \href{https://www.nature.com/articles/s41746-024-01235-0}{paper} /
  \href{https://github.com/ipolharvard/ethos-ares}{code} \\

\end{longtable}
\endgroup

\twocolumn

\section{Analysis definitions}
\label{app:analysis_details}

\subsection{Statistical analysis overview}
\label{app:statistics}

Our primary quantity of interest is the difference in held-out AUROC between two
\methods on a given task and evaluation setting. We use these pairwise
differences, together with the aggregate rankings across tasks that they induce,
to assess how comparative conclusions change when the dataset or task family
changes. Formal definitions follow in the remainder of this appendix, and
uncertainty quantification is detailed in Appendix~\ref{app:uncertainty}.

\paragraph{Transfer across evaluation settings (Q1).}
We compare the aggregate \method rankings for every pair of the four settings.
Pairwise agreement is the proportion of \method pairs whose mean-AUROC
difference has the same sign in both settings. Agreement above chance is
assessed with one-sided random-ranking permutation tests ($10^6$ permutations,
plus-one correction), with Holm adjustment across the six between-setting
comparisons.

\paragraph{Task--\method interactions (Q2).}
We quantify within-setting task consistency as one minus the task-level
reversal rate: the proportion of task-specific pairwise \method comparisons that
agree with the corresponding comparison averaged across tasks in the same
setting (Appendix~\ref{app:reversal-rate}). We complement this with a
descriptive decomposition of each task--\method AUROC matrix into \method,
task, and \method--task interaction components. Further exploratory Q2
analyses are described alongside their results in Section~\ref{sec:results};
where significance is reported for these analyses, it is assessed with
permutation tests, Holm-adjusted across datasets where applicable. Q3 relates
aggregate performance to model-family artifact dates, with component-precedent
and performance--compute analyses reported in Appendices~\ref{app:model_age}
and~\ref{app:compute_tradeoff}.

\subsection{AUROC and pairwise \method differences}
\label{app:auc}

For a binary prediction task, let $S^{+}$ and $S^{-}$ denote the scores
assigned by a model to randomly sampled positive and negative examples,
respectively. The area under the receiver operating characteristic curve
(AUROC) can be interpreted as the probability that a randomly selected
positive example receives a higher score than a randomly selected negative
example:
\[
\mathrm{AUROC}
=
\Pr(S^{+} > S^{-})
+
\frac{1}{2}\Pr(S^{+}=S^{-}).
\]

For \method $m$, task $t$, and evaluation setting $s$, we denote the held-out
AUROC by $A_{mts}$. Our primary comparative quantity is the difference in
AUROC between two \methods $i$ and $j$:
\[
\Delta_{ijts}
=
A_{its}-A_{jts}.
\]
Positive values indicate that \method $i$ outperforms \method $j$ on task $t$.

When comparing \methods at the evaluation-setting level, we first average AUROC
across the $T_s$ tasks in that setting:
\[
\bar A_{ms}
=
\frac{1}{T_s}\sum_{t=1}^{T_s} A_{mts},
\]
and define the corresponding aggregate pairwise difference as
\[
\bar\Delta_{ijs}
=
\bar A_{is}-\bar A_{js}.
\]

\subsection{Task-level reversal rate}
\label{app:reversal-rate}

We use task-level reversals to quantify how often an individual task favors the
opposite \method from the aggregate comparison within the same evaluation
setting. A reversal occurs for \method pair $(i,j)$ on task $t$ when the
task-specific difference and aggregate difference have opposite signs:
\[
\Delta_{ijts}\,\bar\Delta_{ijs} < 0.
\]

The reversal rate for evaluation setting $s$ is therefore
\[
R_s
=
\frac{1}{T_s\binom{M}{2}}
\sum_{t=1}^{T_s}
\sum_{i<j}
\mathbf{1}
\left[
\Delta_{ijts}\,\bar\Delta_{ijs} < 0
\right],
\]
where $M$ is the number of evaluated \methods and $\mathbf{1}[\cdot]$ is the
indicator function. Thus, $R_s$ is the proportion of all task--\method-pair
comparisons whose direction disagrees with the corresponding aggregate
comparison. Ties are not counted as reversals.

\subsection{Uncertainty quantification}
\label{app:uncertainty}
For ranking agreement, reversal rates, horizon correlations, and
training-strategy differences, we combine the 2,000 supplied joint patient-bootstrap draws
with task resampling. Within each dataset, a common patient draw is reused
across all tasks and \methods; draw orders are paired independently across
datasets. Independently within each setting, we sample ten tasks with
replacement, identically across \methods, and recompute each statistic and
its aggregate references. For horizon correlations, each sampled task's
horizon and contrast remain paired and ranks are recomputed; undefined
correlations from constant resamples are excluded and counted. We report
marginal 95\% percentile intervals, assuming exchangeable tasks within
settings. These intervals condition on
the fitted \methods: repeated \method training was not bootstrapped because of
computational cost. 

\section{Dataset characterization}
\label{app:datasets}

Dataset characterization is shown in Table \ref{tab:datasets}.

\begin{table}[t]
\centering
\small
\caption{Summary of the two datasets used in our experiments.}
\label{tab:datasets}
\begin{tabular}{lrr}
\toprule
\textbf{Characteristic} & \textbf{MIMIC-IV} & \textbf{NWICU} \\
\midrule
Subjects & 364,627 & 25,923 \\
Events & 881M & 50.8M \\
Events / subject & 2,417 & 1,959 \\
Vocabulary size & 148,193 & 26,620 \\
\shortstack[l]{Subjects with\\ICU stay} & 65,366 (17.9\%) & 23,204 (89.5\%) \\
\bottomrule
\end{tabular}
\end{table}

\section{Tasks}
\label{app:tasks}

\subsection{Task specifications}
\label{app:tasks_labels}

Task specifications are shown in Tables \ref{tab:mimic-generated-tasks}, \ref{tab:nwicu-generated-tasks}, \ref{tab:mimic-clinical-tasks}, \ref{tab:nwicu-clinical-tasks}.

\begin{table*}[t]
\centering
\footnotesize
\setlength{\tabcolsep}{4pt}
\caption{\Gentasks tasks selected on MIMIC-IV. All statistics are measured
on the held-out split. The \gentasks task selection requires
$\min(n_{\mathrm{pos}}, n_{\mathrm{neg}}) \geq 150$. $\Delta t$ is the
prediction horizon; Prev.\ is the prevalence, with the minority-class count
$n_{\min}$ in parentheses. Avg AUROC is averaged across the 12 models.}
\label{tab:mimic-generated-tasks}
\begin{tabular}{@{}llllrr@{}}
\toprule
\multicolumn{2}{@{}l}{\textbf{Task}} & \multirow{2}{*}{\textbf{Category}} & \multirow{2}{*}{$\boldsymbol{\Delta t}$} & \multirow{2}{*}{\textbf{Prev.\ ($n_{\min}$)}} & \multirow{2}{*}{\textbf{Avg AUROC}} \\
\cmidrule(r){1-2}
\textbf{ID} & \textbf{Name} & & & & \\
\midrule
MG3 & MCH result & LAB & 1\,d & 51.13\% (11,235) & 0.854 \\
MG7 & Death & MEDS\_DEATH & 1\,d & 1.01\% (233) & 0.910 \\
MG8 & IV acetaminophen & INFUSION\_START & 2\,d & 2.97\% (644) & 0.919 \\
MG5 & 22-gauge IV catheter removal & PROCEDURE & 2\,d & 2.04\% (443) & 0.904 \\
MG6 & Oral-gastric output & SUBJECT\_FLUID\_OUTPUT & 2\,d & 1.12\% (244) & 0.909 \\
MG9 & Invasive ventilation end & PROCEDURE & 30\,d & 5.44\% (946) & 0.926 \\
MG10 & Tube-feed residual & SUBJECT\_FLUID\_OUTPUT & 30\,d & 2.75\% (477) & 0.926 \\
MG2 & Serum chloride & LAB & 180\,d & 70.92\% (3,954) & 0.839 \\
MG4 & Phenylephrine infusion & INFUSION\_START & 180\,d & 2.64\% (362) & 0.883 \\
MG1 & 20-gauge IV catheter placement & PROCEDURE & 365\,d & 8.99\% (872) & 0.812 \\
\bottomrule
\end{tabular}
\end{table*}


\begin{table*}[t]
\centering
\footnotesize
\setlength{\tabcolsep}{4pt}
\caption{\Gentasks tasks selected on NWICU. All statistics are measured
on the held-out split. The \gentasks task selection requires
$\min(n_{\mathrm{pos}}, n_{\mathrm{neg}}) \geq 150$. Columns as in
Table~\ref{tab:mimic-generated-tasks}.}
\label{tab:nwicu-generated-tasks}
\begin{tabular}{@{}llllrr@{}}
\toprule
\multicolumn{2}{@{}l}{\textbf{Task}} & \multirow{2}{*}{\textbf{Category}} & \multirow{2}{*}{$\boldsymbol{\Delta t}$} & \multirow{2}{*}{\textbf{Prev.\ ($n_{\min}$)}} & \multirow{2}{*}{\textbf{Avg AUROC}} \\
\cmidrule(r){1-2}
\textbf{ID} & \textbf{Name} & & & & \\
\midrule
NG7 & Diet order & PROCEDURE & 3\,d & 14.5\% (348) & 0.792 \\
NG5 & Heparin injection & MEDICATION & 3\,d & 8.9\% (214) & 0.742 \\
NG6 & Total bilirubin & LAB & 7\,d & 41.6\% (866) & 0.763 \\
NG8 & Patient transfer & PROCEDURE & 7\,d & 25.3\% (522) & 0.807 \\
NG4 & Magnesium sulfate IV & MEDICATION & 14\,d & 22.8\% (323) & 0.729 \\
NG3 & Potassium chloride, oral & MEDICATION & 14\,d & 22.8\% (323) & 0.704 \\
NG2 & Immature granulocytes & LAB & 30\,d & 44.4\% (490) & 0.703 \\
NG9 & Death & MEDS\_DEATH & 90\,d & 36.5\% (318) & 0.809 \\
NG10 & Pulse oximetry & LAB & 180\,d & 38.8\% (294) & 0.819 \\
NG1 & Portable chest X-ray & PROCEDURE & 731\,d & 32.5\% (163) & 0.688 \\
\bottomrule
\end{tabular}
\end{table*}


\begin{table*}[t]
\centering
\footnotesize
\setlength{\tabcolsep}{4pt}
\caption{\Clintasks tasks evaluated on MIMIC-IV. All statistics are measured
on the held-out split. Clinical tasks are retained when
$\min(n_{\mathrm{pos}}, n_{\mathrm{neg}}) \geq 40$. Columns as in
Table~\ref{tab:mimic-generated-tasks}.}
\label{tab:mimic-clinical-tasks}
\begin{tabular}{@{}llllrr@{}}
\toprule
\multicolumn{2}{@{}l}{\textbf{Task}} & \multirow{2}{*}{\textbf{Anchor}} & \multirow{2}{*}{$\boldsymbol{\Delta t}$} & \multirow{2}{*}{\textbf{Prev.\ ($n_{\min}$)}} & \multirow{2}{*}{\textbf{Avg AUROC}} \\
\cmidrule(r){1-2}
\textbf{ID} & \textbf{Name} & & & & \\
\midrule
MC8 & Circulatory failure & ICU admission + 24 h & 8\,h  & 0.86\% (45) & 0.792 \\
MC4 & Invasive ventilation & ICU admission + 24 h & 12\,h & 1.16\% (69) & 0.752 \\
MC9 & Thrombocytopenia & Hospital admission + 24 h & 24\,h & 5.17\% (765) & 0.825 \\
MC7 & Respiratory failure & ICU admission + 24 h & 24\,h & 4.83\% (226) & 0.788 \\
MC3 & Hyponatremia & Hospital admission + 24 h & 48\,h & 9.38\% (1,065) & 0.737 \\
MC1 & Leukocytosis & ICU admission + 24 h & 48\,h & 13.39\% (464) & 0.700 \\
MC6 & Acute kidney injury & ICU admission + 24 h & 48\,h & 9.00\% (547) & 0.763 \\
MC5 & Hospital discharge & Hospital admission + 48 h & 7\,d & 70.69\% (9,714) & 0.755 \\
MC2 & Hospital readmission & Live hospital discharge & 30\,d & 20.85\% (11,283) & 0.722 \\
MC10 & Post-discharge mortality & Live hospital discharge & 90\,d & 5.09\% (2,755) & 0.876 \\
\bottomrule
\end{tabular}
\end{table*}


\begin{table*}[t]
\centering
\footnotesize
\setlength{\tabcolsep}{4pt}
\caption{\Clintasks tasks evaluated on NWICU. All statistics are measured
on the held-out split. Clinical tasks are retained when
$\min(n_{\mathrm{pos}}, n_{\mathrm{neg}}) \geq 40$. Columns as in
Table~\ref{tab:mimic-generated-tasks}.}
\label{tab:nwicu-clinical-tasks}
\begin{tabular}{@{}llllrr@{}}
\toprule
\multicolumn{2}{@{}l}{\textbf{Task}} & \multirow{2}{*}{\textbf{Anchor}} & \multirow{2}{*}{$\boldsymbol{\Delta t}$} & \multirow{2}{*}{\textbf{Prev.\ ($n_{\min}$)}} & \multirow{2}{*}{\textbf{Avg AUROC}} \\
\cmidrule(r){1-2}
\textbf{ID} & \textbf{Name} & & & & \\
\midrule
NC9 & Thrombocytopenia & Hospital admission + 24 h & 24\,h & 7.20\% (157) & 0.746 \\
NC1 & Leukocytosis & ICU admission + 24 h & 48\,h & 16.27\% (192) & 0.670 \\
NC3 & Hyponatremia & Hospital admission + 24 h & 48\,h & 9.28\% (164) & 0.699 \\
NC4 & Sedation initiation & ICU admission + 24 h & 48\,h & 6.37\% (118) & 0.702 \\
NC5 & Broad-spectrum antibiotic initiation & ICU admission + 24 h & 48\,h & 4.47\% (86) & 0.703 \\
NC6 & Vasopressor initiation & ICU admission + 24 h & 48\,h & 4.89\% (83) & 0.714 \\
NC8 & Acute kidney injury & ICU admission + 24 h & 48\,h & 4.14\% (79) & 0.744 \\
NC7 & Hospital discharge & Hospital admission + 48 h & 7\,d & 59.94\% (1,806) & 0.721 \\
NC2 & Hospital readmission & Live hospital discharge & 30\,d & 29.18\% (1,635) & 0.672 \\
NC10 & Post-discharge mortality & Live hospital discharge & 90\,d & 6.67\% (374) & 0.748 \\
\bottomrule
\end{tabular}
\end{table*}

\subsection{Task adequacy threshold.}
We impose a minimum cohort size to avoid evaluating models on tasks for which
the held-out AUROC is estimated with substantial sampling uncertainty. AUROC
can be interpreted as the probability that a randomly selected positive
example is ranked above a randomly selected negative example
\citep{hanley1982meaning}. For a Bernoulli ranking outcome with probability
$A$, the variance is $A(1-A)$, which is maximized at $A=0.5$ and is therefore
bounded by $1/4$. This motivates the simple resolution heuristic
\[
    \mathrm{SE}(\mathrm{AUROC})
    \lesssim \frac{1}{2\sqrt{N}},
    \qquad
    N = \min(n_{\mathrm{pos}},n_{\mathrm{neg}}),
\]
where the minority-class size is used because it is the limiting class for
estimating pairwise discrimination.

Under this heuristic, $N=150$ corresponds to an AUROC standard-error scale of
approximately $0.041$, while $N=40$ corresponds to approximately $0.079$.
We use these values as task-selection criteria rather than as exact confidence
intervals; the actual sampling variance of AUROC depends on both class sizes
and the distribution of model scores.
\label{app:task-floor}

\subsection{Matched tasks}
\label{app:matched_tasks}

Matched tasks are summarized in Table \ref{tab:q2h-matched-tasks}.
\begin{table}[t]
\centering
\small
\setlength{\tabcolsep}{4pt}
\caption{Clinical task pairs treated as conceptually matched in the cross-dataset ranking-similarity analysis.}
\label{tab:q2h-matched-tasks}
\begin{tabular}{@{}lll@{}}
\toprule
Clinical concept & MIMIC-IV & NWICU \\
\midrule
Hyponatremia & MC3 & NC3 \\
Thrombocytopenia & MC9 & NC9 \\
Leukocytosis & MC1 & NC1 \\
Hospital readmission (30 d) & MC2 & NC2 \\
Post-discharge mortality (90 d) & MC10 & NC10 \\
Length of stay (7 d) & MC5 & NC7 \\
\bottomrule
\end{tabular}

\end{table}

\section{Model development}
\label{app:model_development}

We used Claude Code v2.1 with Anthropic Claude Opus 5, assigning one agent to each model implementation. The agents were not given access to raw PhysioNet data. Experiments were run on a compute cluster with NVIDIA L40S GPUs.

\section{Detailed results}
\label{app:detailed_results}

Detailed AUROC results are shown in Tables \ref{tab:appendix-performance-mg}, \ref{tab:appendix-performance-mc}, \ref{tab:appendix-performance-nc}, \ref{tab:appendix-performance-ng}.

Detailed compute cost results are shown in Tables \ref{tab:q3c-compute-cost-mimic}, \ref{tab:q3c-compute-cost-nwicu} and \ref{tab:q3c-compute-cost-pooled}.

\begin{table*}[t]
\centering
\small
\caption{Task-level AUROC for MIMIC-IV Generated (MG). Mean is the unweighted task average; bold indicates the highest unrounded value in each column. Point estimates only.}
\label{tab:appendix-performance-mg}
\setlength{\tabcolsep}{3pt}
\begin{tabular}{l*{11}{r}}
\toprule
Model & MG1 & MG2 & MG3 & MG4 & MG5 & MG6 & MG7 & MG8 & MG9 & MG10 & Mean \\
\midrule
MEDS-TAB-large & 0.8697 & 0.9075 & \textbf{0.9344} & 0.9470 & 0.9695 & \textbf{0.9779} & \textbf{0.9804} & 0.9768 & 0.9782 & 0.9725 & \textbf{0.9514} \\
MEDS-EIC-AR & 0.8500 & 0.9070 & 0.9325 & 0.9357 & 0.9675 & 0.9699 & 0.9624 & 0.9708 & 0.9718 & 0.9700 & 0.9438 \\
MOTOR-FT & \textbf{0.8703} & \textbf{0.9107} & 0.9276 & \textbf{0.9540} & \textbf{0.9712} & 0.9740 & 0.9735 & \textbf{0.9783} & \textbf{0.9801} & 0.9742 & 0.9514 \\
MEDS-EIC-AR-sup & 0.8268 & 0.8842 & 0.9051 & 0.9219 & 0.9541 & 0.9594 & 0.9538 & 0.9638 & 0.9630 & 0.9661 & 0.9298 \\
MOTOR-LP & 0.8700 & 0.8951 & 0.8950 & 0.9499 & 0.9388 & 0.9720 & 0.9707 & 0.9781 & 0.9796 & \textbf{0.9746} & 0.9424 \\
ICU-XGBoost & 0.8256 & 0.7936 & 0.8079 & 0.9134 & 0.9165 & 0.9596 & 0.9626 & 0.9401 & 0.9584 & 0.9632 & 0.9041 \\
RETAIN & 0.8361 & 0.8838 & 0.9097 & 0.9125 & 0.9529 & 0.9511 & 0.9328 & 0.9548 & 0.9575 & 0.9637 & 0.9255 \\
DuETT & 0.8259 & 0.8285 & 0.8604 & 0.9155 & 0.9247 & 0.9420 & 0.9581 & 0.9441 & 0.9462 & 0.9557 & 0.9101 \\
MEDS-TAB-tiny & 0.8065 & 0.7826 & 0.8182 & 0.8845 & 0.9010 & 0.9268 & 0.9366 & 0.9241 & 0.9357 & 0.9605 & 0.8877 \\
TECO & 0.7922 & 0.7682 & 0.7716 & 0.8750 & 0.9071 & 0.9270 & 0.9181 & 0.9109 & 0.9297 & 0.9520 & 0.8752 \\
MEDS-TAB-LR & 0.7393 & 0.8712 & 0.9023 & 0.7535 & 0.8638 & 0.7815 & 0.7598 & 0.8898 & 0.8974 & 0.8386 & 0.8297 \\
BEHRT & 0.6343 & 0.6376 & 0.5855 & 0.6295 & 0.5844 & 0.5722 & 0.6115 & 0.5983 & 0.6148 & 0.6223 & 0.6090 \\
\bottomrule
\end{tabular}
\end{table*}

\begin{table*}[t]
\centering
\small
\caption{Task-level AUROC for MIMIC-IV Clinical (MC). Mean is the unweighted task average; bold indicates the highest unrounded value in each column. Point estimates only.}
\label{tab:appendix-performance-mc}
\setlength{\tabcolsep}{3pt}
\begin{tabular}{l*{11}{r}}
\toprule
Model & MC1 & MC2 & MC3 & MC4 & MC5 & MC6 & MC7 & MC8 & MC9 & MC10 & Mean \\
\midrule
MEDS-TAB-large & \textbf{0.8105} & \textbf{0.7847} & \textbf{0.8090} & \textbf{0.8759} & \textbf{0.8187} & \textbf{0.8579} & \textbf{0.8935} & \textbf{0.9006} & \textbf{0.9288} & \textbf{0.9225} & \textbf{0.8602} \\
MEDS-EIC-AR & 0.7334 & 0.7591 & 0.7991 & 0.8196 & 0.8121 & 0.8363 & 0.8682 & 0.8848 & 0.9063 & 0.9103 & 0.8329 \\
MOTOR-FT & 0.7625 & 0.7598 & 0.8017 & 0.8376 & 0.8082 & 0.7974 & 0.8481 & 0.8798 & 0.9123 & 0.9095 & 0.8317 \\
MEDS-EIC-AR-sup & 0.6923 & 0.7476 & 0.7766 & 0.8050 & 0.7980 & 0.8248 & 0.8473 & 0.8466 & 0.8783 & 0.9051 & 0.8121 \\
MOTOR-LP & 0.7104 & 0.7550 & 0.7365 & 0.8389 & 0.7993 & 0.7952 & 0.8542 & 0.8783 & 0.8070 & 0.9070 & 0.8082 \\
ICU-XGBoost & 0.7755 & 0.6551 & 0.7900 & 0.8266 & 0.7283 & 0.7981 & 0.8474 & 0.8817 & 0.9070 & 0.8388 & 0.8049 \\
RETAIN & 0.6991 & 0.7359 & 0.6992 & 0.7388 & 0.7520 & 0.7491 & 0.8459 & 0.7404 & 0.7513 & 0.8778 & 0.7589 \\
DuETT & 0.7538 & 0.7056 & 0.7918 & 0.7029 & 0.7454 & 0.7571 & 0.7135 & 0.8575 & 0.9075 & 0.8544 & 0.7789 \\
MEDS-TAB-tiny & 0.6903 & 0.6878 & 0.7685 & 0.7362 & 0.7253 & 0.7483 & 0.7785 & 0.7697 & 0.8930 & 0.8368 & 0.7634 \\
TECO & 0.6054 & 0.6391 & 0.6523 & 0.7100 & 0.6954 & 0.7242 & 0.7762 & 0.7947 & 0.6863 & 0.7846 & 0.7068 \\
MEDS-TAB-LR & 0.6129 & 0.7179 & 0.6495 & 0.6103 & 0.7647 & 0.6853 & 0.6409 & 0.6457 & 0.6960 & 0.8641 & 0.6887 \\
BEHRT & 0.5524 & 0.7223 & 0.5664 & 0.5211 & 0.6076 & 0.5771 & 0.5450 & 0.4243 & 0.6220 & 0.9069 & 0.6045 \\
\bottomrule
\end{tabular}
\end{table*}

\begin{table*}[t]
\centering
\small
\caption{Task-level AUROC for NWICU Clinical (NC). Mean is the unweighted task average; bold indicates the highest unrounded value in each column. Point estimates only.}
\label{tab:appendix-performance-nc}
\setlength{\tabcolsep}{3pt}
\begin{tabular}{l*{11}{r}}
\toprule
Model & NC1 & NC2 & NC3 & NC4 & NC5 & NC6 & NC7 & NC8 & NC9 & NC10 & Mean \\
\midrule
MEDS-TAB-large & \textbf{0.7847} & \textbf{0.7700} & \textbf{0.8156} & \textbf{0.8012} & \textbf{0.7957} & \textbf{0.8300} & \textbf{0.8128} & 0.8437 & 0.8983 & \textbf{0.8279} & \textbf{0.8180} \\
MEDS-EIC-AR & 0.7534 & 0.7425 & 0.7826 & 0.7527 & 0.7946 & 0.8237 & 0.8105 & 0.8046 & 0.8898 & 0.8146 & 0.7969 \\
MOTOR-FT & 0.6186 & 0.7394 & 0.6523 & 0.7737 & 0.7388 & 0.7524 & 0.7456 & 0.7610 & 0.6921 & 0.7548 & 0.7229 \\
MEDS-EIC-AR-sup & 0.6728 & 0.7288 & 0.7127 & 0.7356 & 0.7577 & 0.7683 & 0.7569 & 0.7486 & 0.7300 & 0.7973 & 0.7408 \\
MOTOR-LP & 0.6317 & 0.7052 & 0.6258 & 0.7578 & 0.7656 & 0.7647 & 0.7191 & 0.7664 & 0.6907 & 0.7434 & 0.7170 \\
ICU-XGBoost & 0.7706 & 0.6163 & 0.8131 & 0.6902 & 0.7186 & 0.7267 & 0.7150 & \textbf{0.8589} & \textbf{0.9016} & 0.7775 & 0.7589 \\
RETAIN & 0.6461 & 0.6797 & 0.6791 & 0.7540 & 0.6892 & 0.7212 & 0.7678 & 0.7512 & 0.7017 & 0.7630 & 0.7153 \\
DuETT & 0.7177 & 0.6193 & 0.7490 & 0.6888 & 0.7192 & 0.7457 & 0.6890 & 0.7901 & 0.8542 & 0.7662 & 0.7339 \\
MEDS-TAB-tiny & 0.6782 & 0.6396 & 0.7422 & 0.7006 & 0.7179 & 0.7329 & 0.6972 & 0.7676 & 0.7661 & 0.7459 & 0.7188 \\
TECO & 0.6059 & 0.5998 & 0.6051 & 0.6617 & 0.6677 & 0.6927 & 0.6892 & 0.7838 & 0.6190 & 0.6820 & 0.6607 \\
MEDS-TAB-LR & 0.5634 & 0.6195 & 0.6329 & 0.6138 & 0.5342 & 0.5219 & 0.6878 & 0.5518 & 0.6254 & 0.5962 & 0.5947 \\
BEHRT & 0.5951 & 0.6003 & 0.5738 & 0.4902 & 0.5341 & 0.4828 & 0.5593 & 0.5035 & 0.5856 & 0.7097 & 0.5634 \\
\bottomrule
\end{tabular}
\end{table*}

\begin{table*}[t]
\centering
\small
\caption{Task-level AUROC for NWICU Generated (NG). Mean is the unweighted task average; bold indicates the highest unrounded value in each column. Point estimates only.}
\label{tab:appendix-performance-ng}
\setlength{\tabcolsep}{3pt}
\begin{tabular}{l*{11}{r}}
\toprule
Model & NG1 & NG2 & NG3 & NG4 & NG5 & NG6 & NG7 & NG8 & NG9 & NG10 & Mean \\
\midrule
MEDS-TAB-large & \textbf{0.7838} & \textbf{0.8020} & \textbf{0.7879} & 0.7966 & \textbf{0.8589} & \textbf{0.8522} & 0.8764 & \textbf{0.9131} & \textbf{0.8770} & \textbf{0.9021} & \textbf{0.8450} \\
MEDS-EIC-AR & 0.7243 & 0.7591 & 0.7768 & \textbf{0.8049} & 0.8429 & 0.8349 & \textbf{0.8813} & 0.9036 & 0.8549 & 0.8597 & 0.8242 \\
MOTOR-FT & 0.7707 & 0.7295 & 0.7430 & 0.7832 & 0.7741 & 0.7722 & 0.8524 & 0.8627 & 0.8497 & 0.8460 & 0.7983 \\
MEDS-EIC-AR-sup & 0.6885 & 0.7289 & 0.6913 & 0.7504 & 0.7776 & 0.7952 & 0.8039 & 0.8272 & 0.8262 & 0.8468 & 0.7736 \\
MOTOR-LP & 0.7602 & 0.6900 & 0.7264 & 0.7808 & 0.7806 & 0.7476 & 0.8365 & 0.8430 & 0.8441 & 0.8114 & 0.7821 \\
ICU-XGBoost & 0.6570 & 0.6521 & 0.7446 & 0.7239 & 0.7047 & 0.7948 & 0.7641 & 0.7696 & 0.8443 & 0.8352 & 0.7490 \\
RETAIN & 0.7146 & 0.7806 & 0.7371 & 0.7897 & 0.7848 & 0.8219 & 0.8453 & 0.8798 & 0.8367 & 0.8663 & 0.8057 \\
DuETT & 0.6602 & 0.7457 & 0.6989 & 0.7481 & 0.7391 & 0.8040 & 0.8078 & 0.8166 & 0.8426 & 0.8585 & 0.7722 \\
MEDS-TAB-tiny & 0.6505 & 0.7239 & 0.6599 & 0.7120 & 0.7379 & 0.7737 & 0.7115 & 0.7234 & 0.7723 & 0.7965 & 0.7262 \\
TECO & 0.6679 & 0.5897 & 0.6414 & 0.6414 & 0.6818 & 0.6804 & 0.7609 & 0.7676 & 0.7572 & 0.8684 & 0.7057 \\
MEDS-TAB-LR & 0.6291 & 0.6673 & 0.6897 & 0.6594 & 0.6726 & 0.7234 & 0.7488 & 0.7704 & 0.7820 & 0.7635 & 0.7106 \\
BEHRT & 0.5465 & 0.5720 & 0.5488 & 0.5606 & 0.5498 & 0.5605 & 0.6103 & 0.6090 & 0.6187 & 0.5686 & 0.5745 \\
\bottomrule
\end{tabular}
\end{table*}

\begin{table*}[t]
\centering
\small
\setlength{\tabcolsep}{4pt}
\caption{MIMIC-IV generated tasks: macro-AUC and mean pipeline wall time per task for the 8 retained tasks. Preprocessing and pretraining include the allocated shared costs; extraction is required for MOTOR-LP feature extraction. Training denotes fine-tuning or supervised fitting. Total is in hours; components are in seconds.}
\label{tab:q3c-compute-cost-mimic}
\begin{tabular}{lrrrrrr}
\toprule
Model & Macro-AUC & Preprocess (s) & Pretrain (s) & Train (s) & Extract (s) & Total (h) \\
\midrule
MEDS-TAB-large & 0.953529 & 38,776.408 & 0.000 & 62,360.200 & 0.000 & 28.093502 \\
MEDS-EIC-AR & 0.944870 & 2,768.338 & 1,085.950 & 3,422.240 & 0.000 & 2.021258 \\
MOTOR-FT & 0.952875 & 688.006 & 6,210.695 & 230,571.552 & 0.000 & 65.963959 \\
MOTOR-LP & 0.943626 & 688.006 & 6,210.695 & 152.098 & 2,439.183 & 2.636106 \\
RETAIN & 0.926711 & 152.375 & 0.000 & 155.875 & 0.000 & 0.085625 \\
MEDS-EIC-AR-sup & 0.931370 & 2,768.338 & 0.000 & 3,367.908 & 0.000 & 1.704513 \\
DuETT & 0.915796 & 9.800 & 10,830.688 & 874.023 & 0.000 & 3.254031 \\
ICU-XGBoost & 0.911114 & 6.219 & 0.000 & 5.078 & 0.000 & 0.003138 \\
MEDS-TAB-tiny & 0.894786 & 2,561.875 & 0.000 & 4,486.750 & 0.000 & 1.957951 \\
TECO & 0.881755 & 33.087 & 0.000 & 9,693.292 & 0.000 & 2.701772 \\
MEDS-TAB-LR & 0.816080 & 38,776.408 & 0.000 & 604.022 & 0.000 & 10.939008 \\
BEHRT & 0.604747 & 2.625 & 221.780 & 1,301.012 & 0.000 & 0.423727 \\
\bottomrule
\end{tabular}
\par\smallskip
\begin{minipage}{\textwidth}\footnotesize
Totals use unrounded components. Prediction time is excluded. Zero denotes a component not used. Wall times are not hardware-normalized; values describe the retained task set, not the full evaluation.
\end{minipage}
\end{table*}

\begin{table*}[t]
\centering
\small
\setlength{\tabcolsep}{4pt}
\caption{NWICU generated tasks: macro-AUC and mean pipeline wall time per task for the 7 retained tasks. Preprocessing and pretraining include the allocated shared costs; extraction is required for MOTOR-LP feature extraction. Training denotes fine-tuning or supervised fitting. Total is in hours; components are in seconds.}
\label{tab:q3c-compute-cost-nwicu}
\begin{tabular}{lrrrrrr}
\toprule
Model & Macro-AUC & Preprocess (s) & Pretrain (s) & Train (s) & Extract (s) & Total (h) \\
\midrule
MEDS-TAB-large & 0.856071 & 3,688.251 & 0.000 & 6,401.618 & 0.000 & 2.802742 \\
MEDS-EIC-AR & 0.833626 & 212.314 & 255.888 & 242.627 & 0.000 & 0.197452 \\
MOTOR-FT & 0.813803 & 71.783 & 1,128.593 & 23,286.575 & 0.000 & 6.801931 \\
MOTOR-LP & 0.795611 & 71.783 & 1,128.593 & 128.474 & 144.120 & 0.409158 \\
RETAIN & 0.814524 & 1.565 & 0.000 & 9.875 & 0.000 & 0.003178 \\
MEDS-EIC-AR-sup & 0.782722 & 212.314 & 0.000 & 247.324 & 0.000 & 0.127677 \\
DuETT & 0.784100 & 0.879 & 842.774 & 61.184 & 0.000 & 0.251344 \\
ICU-XGBoost & 0.772798 & 0.725 & 0.000 & 0.929 & 0.000 & 0.000459 \\
MEDS-TAB-tiny & 0.726841 & 81.857 & 0.000 & 80.000 & 0.000 & 0.044960 \\
TECO & 0.734837 & 1.766 & 0.000 & 1,034.701 & 0.000 & 0.287907 \\
MEDS-TAB-LR & 0.729546 & 3,688.251 & 0.000 & 38.713 & 0.000 & 1.035268 \\
BEHRT & 0.580357 & 0.169 & 17.906 & 101.415 & 0.000 & 0.033192 \\
\bottomrule
\end{tabular}
\par\smallskip
\begin{minipage}{\textwidth}\footnotesize
Totals use unrounded components. Prediction time is excluded. Zero denotes a component not used. Wall times are not hardware-normalized; values describe the retained task set, not the full evaluation.
\end{minipage}
\end{table*}

\begin{table*}[t]
\centering
\small
\setlength{\tabcolsep}{4pt}
\caption{Pooled generated tasks: macro-AUC and mean pipeline wall time per task for the 15 retained tasks. Preprocessing and pretraining include the allocated shared costs; extraction is required for MOTOR-LP feature extraction. Training denotes fine-tuning or supervised fitting. Total is in hours; components are in seconds. Pooled means give equal weight to each retained task across both datasets.}
\label{tab:q3c-compute-cost-pooled}
\begin{tabular}{lrrrrrr}
\toprule
Model & Macro-AUC & Preprocess (s) & Pretrain (s) & Train (s) & Extract (s) & Total (h) \\
\midrule
MEDS-TAB-large & 0.908048 & 22,401.934 & 0.000 & 36,246.195 & 0.000 & 16.291147 \\
MEDS-EIC-AR & 0.892956 & 1,575.526 & 698.588 & 1,938.421 & 0.000 & 1.170149 \\
MOTOR-FT & 0.887975 & 400.435 & 3,839.048 & 133,838.563 & 0.000 & 38.355013 \\
MOTOR-LP & 0.874552 & 400.435 & 3,839.048 & 141.073 & 1,368.154 & 1.596864 \\
RETAIN & 0.874357 & 81.997 & 0.000 & 87.742 & 0.000 & 0.047150 \\
MEDS-EIC-AR-sup & 0.862001 & 1,575.526 & 0.000 & 1,911.635 & 0.000 & 0.968656 \\
DuETT & 0.854338 & 5.637 & 6,169.661 & 494.698 & 0.000 & 1.852777 \\
ICU-XGBoost & 0.846566 & 3.655 & 0.000 & 3.141 & 0.000 & 0.001888 \\
MEDS-TAB-tiny & 0.816412 & 1,404.533 & 0.000 & 2,430.267 & 0.000 & 1.065222 \\
TECO & 0.813193 & 18.471 & 0.000 & 5,652.616 & 0.000 & 1.575302 \\
MEDS-TAB-LR & 0.775697 & 22,401.934 & 0.000 & 340.211 & 0.000 & 6.317263 \\
BEHRT & 0.593365 & 1.479 & 126.639 & 741.200 & 0.000 & 0.241477 \\
\bottomrule
\end{tabular}
\par\smallskip
\begin{minipage}{\textwidth}\footnotesize
Totals use unrounded components. Prediction time is excluded. Zero denotes a component not used. Wall times are not hardware-normalized; values describe the retained task set, not the full evaluation.
\end{minipage}
\end{table*}

\section{Model recency and methodological progress}
\label{app:model_age}


To assess whether newer \methods correspond to improved performance, we relate
overall mean AUROC to two notions of model ``age": the first documented public
artifact of each model family and the precedent date of selected architectural
or learning components, as shown in Table \ref{tab:q3-model-dates}. For each timeline, we also report the best performance
achieved up to each year within the evaluated model set. Component dates are
editorial choices, and these curves should be interpreted as retrospective
summaries of the models studied here rather than reconstructions of the
historical state of the art.

We consider two notions of recency. First, we date
each model family by the year of its first public artifact, that can be a paper or a GitHub repository
(Figure~\ref{fig:q3-artifact-progress} in the main text). Under this view, there is some evidence of
improvement over time, but the relationship is far from monotonic: several
recent models perform below substantially older approaches, and the strongest
overall performer is the XGBoost-based MEDS-Tab-Large baseline.

\begin{figure}[t]
  \centering
  \includegraphics[width=\columnwidth]{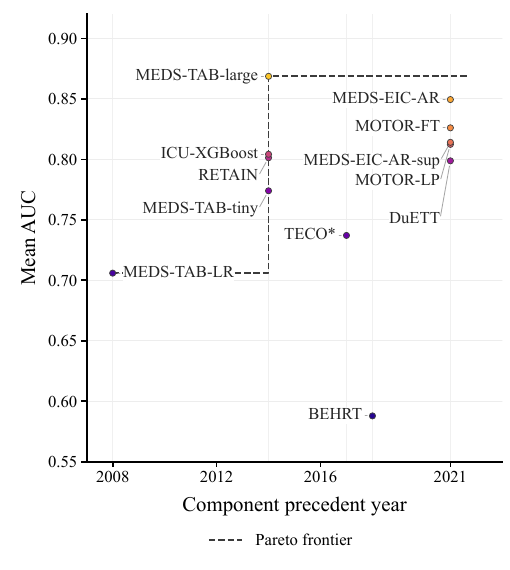}
  \caption{Retrospective progress by selected component-precedent year.
  Points show mean AUROC across the same 40 equally weighted tasks as
  Figure~\ref{fig:q3-artifact-progress}, using identical model colors and
  AUROC limits. The dashed step curve marks the best mean AUROC among evaluated
  models assigned to that year or earlier. Component years are editorial
  precedents (Table~\ref{tab:q3-model-dates}) and do not establish when the
  complete pipelines first became feasible. TECO's asterisk marks a provisional
  Transformer-backbone date.}
  \label{fig:q3-component-progress}
\end{figure}

A different picture emerges when models are dated by the precedent of their
main technical components (Figure~\ref{fig:q3-component-progress}). Strong performance is
already achieved by relatively old modeling components, most notably logistic
regression and XGBoost, and newer technical components do not consistently
improve upon this frontier. This suggests that model recency alone is a poor
proxy for methodological progress in the settings we evaluate.

MEDS-Tab Large illustrates an important distinction between these two views.
Although it is a recent model artifact, its predictive model is XGBoost; its
strong performance instead relies on a modern, extremely wide and sparse
tabularization of the longitudinal record, rather than on a recently introduced
learning algorithm. Thus, some improvements associated with newer systems may
come from how clinical data are represented and exposed to the model, rather
than from the recency of the predictive architecture itself.

\begin{table*}[t]
\centering
\small
\setlength{\tabcolsep}{4pt}
\caption{Model-family artifact dates and selected technical-component dates.}
\label{tab:q3-model-dates}
\begin{tabular}{@{}llll>{\raggedright\arraybackslash}p{0.36\textwidth}@{}}
\toprule
Model name & Artifact date & Artifact type & Technical date & Newest component \\
\midrule
RETAIN & 2016-08-19 & arXiv preprint & 2014 & GRU and attention \\
ICU-XGBoost & 2017 & Conference paper & 2014 & XGBoost \\
BEHRT & 2019-07-22 & arXiv preprint & 2018 & BERT \\
MOTOR-FT & 2023-01-09 & arXiv preprint & 2021 & Rotary positional embeddings \\
MOTOR-LP & 2023-01-09 & arXiv preprint & 2021 & Rotary positional embeddings \\
DuETT & 2023-04-25 & arXiv preprint & 2021 & Continuous-value embeddings (STraTS) \\
MEDS-TAB-LR & 2024-06-05 & Package release & 2008 & Regularized logistic regression \\
MEDS-TAB-tiny & 2024-06-05 & Package release & 2014 & XGBoost \\
MEDS-TAB-large & 2024-06-05 & Package release & 2014 & XGBoost \\
TECO & 2025-01-23 & medRxiv preprint & 2017 & Transformer with sinusoidal positional encoding \\
MEDS-EIC-AR & 2025-04-21 & Repository release & 2021 & Rotary embeddings and parallel residuals \\
MEDS-EIC-AR-sup & 2025-04-21 & Repository release & 2021 & Rotary embeddings and parallel residuals \\
\bottomrule
\end{tabular}
\end{table*}

\section{Performance--compute tradeoff}
\label{app:compute_tradeoff}

\begin{figure*}[t]
\centering
\includegraphics[width=\textwidth,clip,trim={0cm 0.5cm 0cm 2cm}]{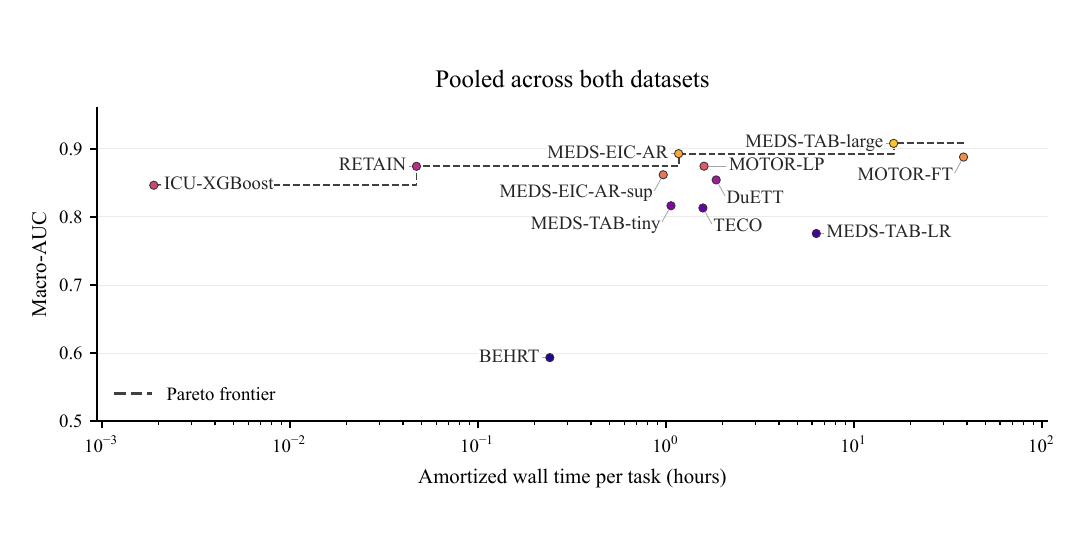}
\caption{Generated-task macro-AUROC versus amortized wall time for 12 models on 8 MIMIC-IV and 7 NWICU tasks. All models use the same retained tasks within each dataset. Shared preprocessing and pretraining are amortized over these tasks; task-specific preprocessing, training, and required feature extraction are added. Dashed curves show the Pareto frontier. Wall times are not hardware-normalized. Pooled values weight each retained task equally. For the compute analysis, we restrict to tasks for which compute measurements were successfully verified, yielding 8 MIMIC-IV and 7 NWICU generated tasks.}
\label{fig:q3c-compute-progress-pooled}
\end{figure*}

To assess whether newer approaches expand the achievable tradeoff between
predictive performance and computational cost, rather than considering
performance alone, Figure~\ref{fig:q3c-compute-progress-pooled} compares
generated-task macro-AUROC with amortized wall time across the evaluated
\methods. Shared preprocessing and pretraining costs are amortized across tasks,
while task-specific preprocessing, training, and required feature extraction are
included; the Pareto frontier marks \methods that achieve higher performance for
a given cost or lower cost for a given performance. The models span a wide range
of tradeoffs: inexpensive historical approaches such as ICU-XGBoost and RETAIN
achieve relatively strong performance at very low cost, while higher-performing
\methods generally require more computation. The frontier is formed by \methods
from several generations and model families rather than by a monotonic sequence
of increasingly recent architectures: MEDS-Tab-large achieves the highest mean
AUROC at the second-highest cost, whereas MEDS-EIC-AR and RETAIN occupy
different points on the frontier at lower computational cost. Per-model costs
are tabulated in Appendix~\ref{app:detailed_results}.

\section{Additional analyses}
\label{app:new_results}

This section provides the detailed numerical results underlying the analyses in
Section~\ref{sec:results}. For Q1, Table~\ref{tab:q1c-agreement} reports
aggregate pairwise agreement and Kendall rank correlation between MIMIC-IV
\gentasks tasks and the other evaluation settings, including bootstrap
confidence intervals and permutation-test results.

For Q2, Tables~\ref{tab:q2a-task-reversals} and
\ref{tab:q2b-variance-decomposition} provide two complementary summaries of
task--\method interaction: the frequency with which individual-task comparisons
reverse the corresponding aggregate comparison, and the decomposition of
observed AUROC variation into model, task, and model--task interaction
components. Table~\ref{tab:q2h-task-pair-similarity} examines whether
conceptually matched clinical tasks induce more similar model rankings across
MIMIC-IV and NWICU than incorrect correspondences among the same six task
concepts; the broader nonmatched comparison is also reported for completeness.

Tables~\ref{tab:q2c-horizon-association} and
\ref{tab:q2e-behrt-absolute-horizon} report the exploratory analyses relating
prediction horizon to model performance. The former measures how the relative
performance of MOTOR-FT and MEDS-EIC-AR changes with horizon, while the latter
relates horizon to BEHRT's absolute AUROC. Finally,
Table~\ref{tab:q2f-training-strategy} reports paired comparisons of
fine-tuning versus linear probing for MOTOR and pretraining versus
supervised-only training for MEDS-EIC-AR across each evaluation setting and
overall.

\begin{table*}[t]
\centering
\caption{Agreement of aggregate model rankings with MIMIC-IV Generated (MG; \textbf{A}).}
\label{tab:q1c-agreement}
\begin{tabular}{lrrr}
\toprule
Target & Agreement [95\% CI] & Kendall $\tau$ [95\% CI] & $p_{\mathrm{Holm}}$ \\
\midrule
MIMIC-IV Clinical (MC; B) & 90.9\% [83.3, 95.5] & 0.818 [0.667, 0.909] & $4.60\times 10^{-5}$ \\
NWICU Clinical (NC; C) & 80.3\% [69.7, 92.4] & 0.606 [0.394, 0.848] & $0.0026$ \\
NWICU Generated (NG; D) & 92.4\% [81.8, 93.9] & 0.848 [0.636, 0.879] & $1.80\times 10^{-5}$ \\
\bottomrule
\end{tabular}
\end{table*}

\begin{table}[!t]
\centering
\caption{Task-level reversals of aggregate model comparisons. Each setting contains $10\times66=660$ task--model-pair comparisons. MG/MC denote MIMIC-IV Generated/Clinical; NG/NC denote NWICU Generated/Clinical.}
\label{tab:q2a-task-reversals}
\begin{tabular}{lr}
\toprule
Setting & Reversal rate [95\% CI] \\
\midrule
MG & 8.3\% [4.5, 10.6] \\
MC & 15.2\% [10.2, 17.7] \\
NC & 16.7\% [10.9, 19.7] \\
NG & 10.9\% [7.7, 15.5] \\
\bottomrule
\end{tabular}
\end{table}

\begin{table}[!t]
\centering
\caption{Two-way decomposition of observed AUROC variation within each evaluation setting. 
MG/MC denote MIMIC-IV Generated/Clinical; NG/NC denote NWICU Generated/Clinical.}
\label{tab:q2b-variance-decomposition}
\begin{tabular}{lrrr}
\toprule
Setting & Model & Task & Interaction \\
\midrule
MG & 78.2\% & 13.6\% & 8.2\% \\
MC & 51.0\% & 25.1\% & 23.9\% \\
NC & 66.0\% & 9.6\% & 24.4\% \\
NG & 62.3\% & 28.7\% & 9.0\% \\
\bottomrule
\end{tabular}
\end{table}

\begin{table}[t]
\centering
\small
\caption{Similarity of individual-task model rankings across datasets. Entries are mean Kendall $\tau_b$ with 95\% patient-and-task bootstrap intervals.}
\label{tab:q2h-task-pair-similarity}
\begin{tabular}{lrr}
\toprule
Task pairs & $n$ & Mean $\tau_b$ [95\% CI] \\
\midrule
Matched & 6 & 0.601 [0.485, 0.712] \\
All nonmatched & 394 & 0.546 [0.480, 0.596] \\
Same-six nonmatched & 30 & 0.437 [0.352, 0.573] \\
\midrule
\multicolumn{2}{l}{Matched $-$ all nonmatched} & 0.055 [-0.059, 0.182] \\
\multicolumn{2}{l}{Matched $-$ same-six nonmatched} & 0.164 [0.025, 0.284] \\
\bottomrule
\end{tabular}
\par\smallskip
\begin{minipage}{\columnwidth}\footnotesize
Exact two-sided correspondence-permutation test within the six concepts: $p=0.0194$ (720 bijections).
\end{minipage}
\end{table}

\begin{table}[!t]
\centering
\caption{Association between generated-task horizon and BEHRT AUROC. Each dataset contains 10 generated tasks. Brackets give 95\% percentile intervals from 2,000 task-only bootstrap replicates; patient uncertainty is unavailable for absolute AUROC. Two-sided permutation tests use 1,000,000 permutations. Reported p-values are Holm-adjusted across the two datasets within this analysis. Positive correlation indicates better absolute performance at longer horizons.}
\label{tab:q2e-behrt-absolute-horizon}
\begin{tabular}{lrr}
\toprule
Dataset & Spearman $\rho$ [95\% CI] & $p_{\mathrm{Holm}}$ \\
\midrule
MIMIC-IV & 0.805 [0.230, 0.967] & $0.0146$ \\
NWICU & -0.135 [-0.810, 0.690] & $0.7105$ \\
\bottomrule
\end{tabular}
\end{table}



\begin{table}[!t]
\centering
\caption{Association of generated-task horizon with the AUROC difference MOTOR-FT minus MEDS-EIC-AR, across 10 tasks per dataset. Positive Spearman $\rho$ indicates a growing relative advantage for MOTOR-FT at longer horizons. Repeated horizons receive average ranks.}
\label{tab:q2c-horizon-association}
\begin{tabular}{lrr}
\toprule
Dataset & Spearman $\rho$ [95\% CI] & $p_{\mathrm{Holm}}$ \\
\midrule
MIMIC-IV & 0.483 [-0.417, 0.949] & $0.1600$ \\
NWICU & 0.795 [0.067, 0.997] & $0.0163$ \\
\bottomrule
\end{tabular}
\end{table}



\begin{table}[!t]
\centering
\small
\caption{Paired training-strategy comparisons.  Entries are mean AUROC differences in percentage points.
MOTOR compares FT minus LP; EIC-AR compares the pretrained variant minus the supervised-only variant.
}
\label{tab:q2f-training-strategy}
\begin{tabular}{lrr}
\toprule
Setting & MOTOR & EIC-AR \\
\midrule
MG & 0.90 [0.16, 1.71] & 1.39 [0.88, 1.97] \\
MC & 2.35 [0.28, 4.85] & 2.08 [0.93, 3.45] \\
NC & 0.58 [-0.89, 1.95] & 5.61 [2.88, 8.88] \\
NG & 1.63 [0.63, 2.66] & 5.06 [3.06, 6.95] \\
\midrule
Overall & 1.37 [0.67, 2.16] & 3.54 [2.61, 4.53] \\
\bottomrule
\end{tabular}


\end{table}

\end{document}